\documentclass{article}
\usepackage{arxiv}
\usepackage[utf8]{inputenc} 
\usepackage[T1]{fontenc}    
\usepackage{hyperref}       
\usepackage{url}            
\usepackage{booktabs}       
\usepackage{amsfonts}       
\usepackage{nicefrac}       
\usepackage{microtype}      
\usepackage{lipsum}
\usepackage{graphicx}
\graphicspath{ {./figure/} }
\usepackage{subcaption}
\usepackage{siunitx}
\usepackage{nccmath}
\usepackage{graphicx,tikz}
\usepackage{comment}

\newcommand{\TOMOGAN}{TomoGAN}

\title{TomoGAN: Low-Dose Synchrotron X-Ray Tomography with Generative Adversarial Networks}

\author{
Zhengchun Liu \\
Data Science and Learning Division\\
Argonne National Laboratory\\
Lemont, IL 60439, USA\\
\texttt{zhengchun.liu@anl.gov} \\
\And
Tekin Bicer \\
Data Science and Learning Division\\
X-ray Science Division\\
Argonne National Laboratory\\
Lemont, IL 60439, USA
\And
Rajkumar Kettimuthu \\
Data Science and Learning Division\\
Argonne National Laboratory\\
Lemont, IL 60439, USA
\And
Doga Gursoy \\
X-ray Science Division\\
Argonne National Laboratory\\
Lemont, IL 60439, USA
\And
Francesco De Carlo \\
X-ray Science Division\\
Argonne National Laboratory\\
Lemont, IL 60439, USA
\And
Ian Foster \\
Data Science and Learning Division\\
Argonne National Laboratory\\
Lemont, IL 60439, USA
}

\begin{document}
\maketitle
\begin{abstract}
Synchrotron-based x-ray tomography is a noninvasive imaging technique that allows for reconstructing the internal structure of materials at high spatial resolutions from tens of micrometers to a few nanometers. In order to resolve sample features at smaller length scales, however, a higher radiation dose is required. Therefore, the limitation on the achievable resolution is set primarily by noise at these length scales. We present \TOMOGAN{}, a denoising technique based on generative adversarial networks, for improving the quality of reconstructed images for low-dose imaging conditions. We evaluate our approach in two photon-budget-limited experimental conditions: (1) sufficient number of low-dose projections (based on Nyquist sampling), and (2) insufficient or limited number of high-dose projections. In both cases the angular sampling is assumed to be isotropic, and the photon budget throughout the experiment is fixed based on the maximum allowable radiation dose on the sample. Evaluation with both simulated and experimental datasets shows that our approach can significantly reduce noise in reconstructed images, improving the structural similarity score of simulation and experimental data from 0.18 to 0.9 and from 0.18 to 0.41, respectively. Furthermore, the quality of the reconstructed images with filtered back projection followed by our denoising approach exceeds that of reconstructions with the simultaneous iterative reconstruction technique, showing the computational superiority of our approach.
\end{abstract}

\section{Introduction}\label{sec:intro}
X-ray computed tomography (CT) is a common noninvasive imaging modality for resolving the internal structure of materials at synchrotrons~\cite{bonse1996x}. 
In CT, 2D projection images of an object are collected at different views of the object around a common axis, and a numerical reconstruction process is applied afterwards to recover the object's morphology in 3D.
Although CT experiments at synchrotrons can collect data at high spatial and temporal resolution, however, in situ or dose-sensitive experiments require short exposure times to capture relevant dynamic phenomena or to avoid sample damage. 
These low-dose (LD) imaging conditions yield noisy measurements that significantly impact the quality of the resulting 3D reconstructions. 
Similar concerns arise when the number of projections is limited to meet speed and/or dose requirements, such as in  conventional lab-based micro-CT systems, where measurements are collected at discrete rotations of the object.
Thus, we want techniques that can map noisy reconstructions (due to fewer projections, lower resolutions, and/or shorter imaging times with noisy measurements) to an approximation of the ideal image.

Much research has been conducted on methods for improving the quality of noisy low-dose images. 
Broadly, these approaches fall into three categories: (i) methods for denoising measurements/raw data, for example, sinograms or projections~\cite{wang2005sinogram, wang2006penalized,manduca2009projection,Anirudh_2018_CVPR}; (ii) advanced iterative reconstruction algorithms, for example, model-based approaches~\cite{vogel1996iterative, beister2012iterative}; and (iii) methods, especially deep convolution neural network based~\cite{Bazrafkan2019DeepLB,8327873,adler2018deep}, for denoising reconstructed images~\cite{ma2011low}. 
For (i) and (iii), various deep learning (DL) approaches have shown great promise in the context of medical CT imaging~\cite{8332971,wolterink2017generative,deep-img,jimaging4110128,2016arXiv160806993H}.
In the context of synchrotron-based CT, for (i), Yang et al.~\cite{Yang2018} used a deep convolutional neural network (CNN) to denoise prereconstruction short-exposure-time projection images from a network trained by a few long-exposure-time (i.e., high-dose) and short-exposure-time (i.e., low-dose) projection pairs.
They achieved a 10-fold increase in signal-to-noise ratio, enabling the reliable tracing of brain structures in low-dose datasets. 
For (iii), Pelt et al.~\cite{jimaging4110128} trained a mixed-scale dense convolutional neural network\cite{2016arXiv160806993H} in a supervised fashion to learn a mapping from low-dose to normal-dose reconstructions. 
They achieved impressive results on simulation datasets; but unfortunately the performance of proposed model is not thoroughly evaluated on different experimental datasets.

DL techniques use multi-layer (``deep'') neural networks (DNNs) to learn representations of data with multiple levels of abstraction.  
These techniques can discover intricate structure in a dataset by using a back-propagation algorithm to set the internal parameters that are used to transform data as they flow between network layers. 
Recent advances in DL, such as convolutional networks~\cite{LeCun2015}, rectifier linear units (ReLUs)~\cite{AlexNet}, batch normalization~\cite{BatchNorm}, dropout~\cite{Dropout}, and residual learning~\cite{residual}, have enabled exciting new applications in many areas.
DL techniques have been applied successfully to a range of  scientific imaging problems, such as denoising, super-resolution, and image enhancement and restoration~\cite{sr-survey-19,2017arXiv171110925U,srgan,pix2pix2017}.  

\begin{figure}[htb] 
\centering
\begin{subfigure}[h]{0.49\linewidth }
\centering
\includegraphics[width=\textwidth,trim={.7cm 0cm 0cm 0cm},clip]{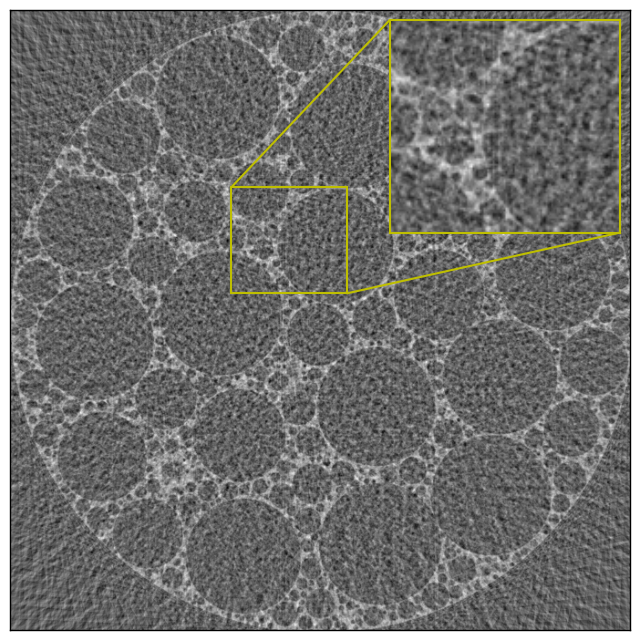}
\caption*{Image produced by conventional reconstruction.}
\end{subfigure}
\begin{subfigure}[h]{0.49\linewidth}
\centering
\includegraphics[width=\textwidth,trim={.7cm 0cm 0cm 0cm},clip]{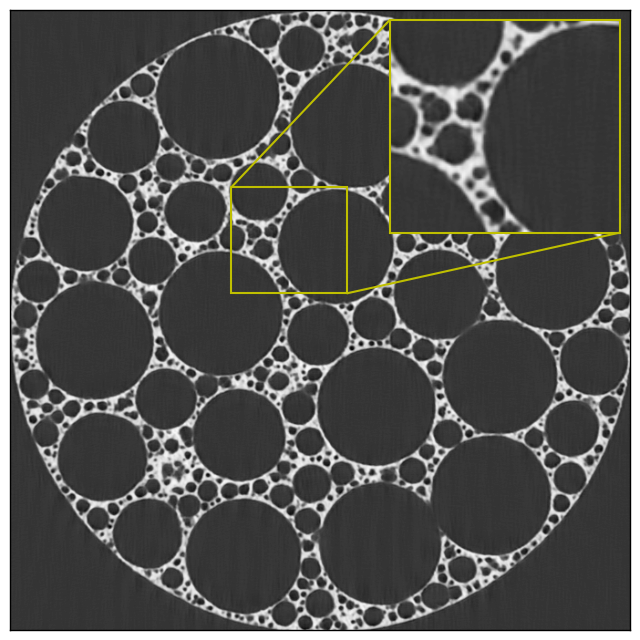}
\caption*{Image following enhancement by \TOMOGAN{}.}
\end{subfigure}
\caption{Two different reconstructions of a noisy simulated dataset, constructed by  subsampling 64 projections from a \num{1024}-projection simulated dataset containing foam features in a 3D volume.
On the left, the results of conventional reconstruction, which are highly noisy.
On the right, those same results after denoising with \TOMOGAN{}; the features are much more visible.
In these images and others that follow, an inset shows details of a representative feature.}
\label{fig:teaser-simu}
\end{figure}

\begin{figure}[htb]
\centering
\begin{subfigure}[h]{0.49\linewidth}
\centering
\includegraphics[width=\textwidth,trim={0cm 3cm 0cm 0cm},clip]{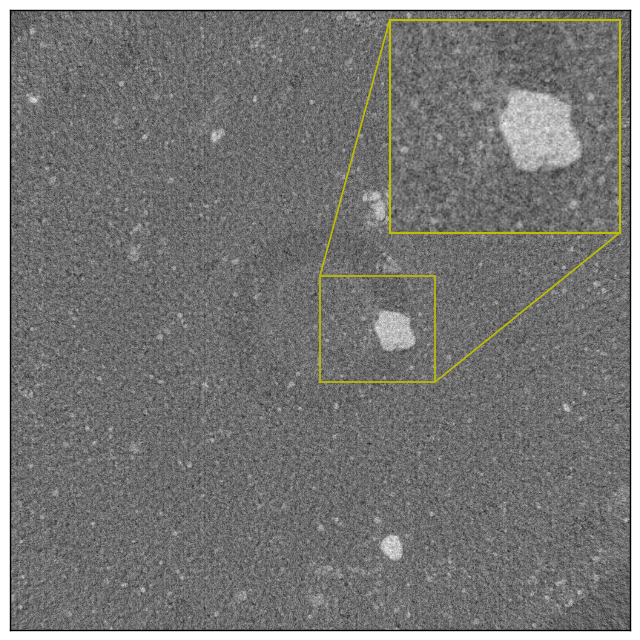}
\caption*{Image produced by conventional reconstruction.}
\end{subfigure}
\begin{subfigure}[h]{0.49\linewidth}
\centering
\includegraphics[width=\textwidth,trim={0cm 3cm 0cm 0cm},clip]{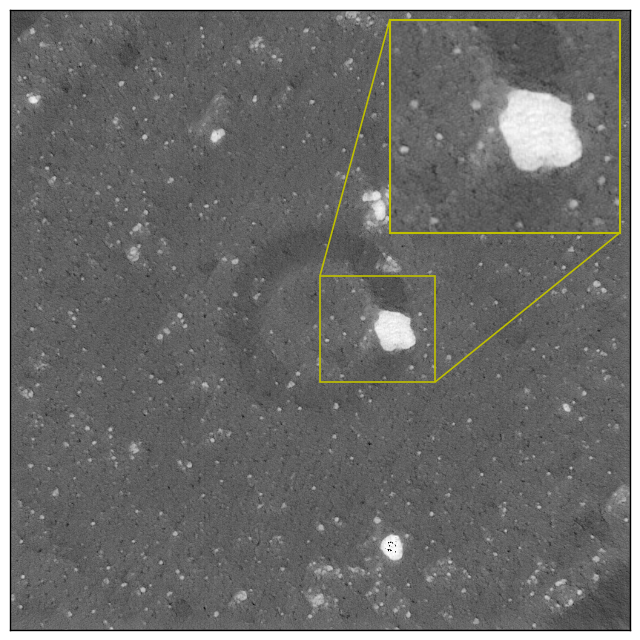}
\caption*{Image following enhancement by \TOMOGAN{}.}
\end{subfigure}
\caption{Two different reconstructions of a noisy experimental dataset,
constructed by subsampling 64 projections from a \num{1024}-projection shale sample dataset.
On the left, the results of conventional reconstruction, which are highly noisy.
On the right, those same results after denoising with \TOMOGAN{}; the features are much more visible.}
\label{fig:teaser-real}
\end{figure}

In this article we explore an alternative DL approach to image enhancement, namely, the use of generative adversarial networks (GANs).
In general, a GAN involves two neural networks, a generator $G$ and a discriminator $D$, that contest with each other in a zero-sum-game framework\cite{2014arXiv1406.2661G}. 
Training a GAN model involves a minimax game between the $G$ that mimics the true distribution and the $D$ that distinguishes samples produced by the $G$ from the real samples. 
Compare with the DNN based methods, the discriminator, provides adversarial loss, is a helper for training the generator to generate better perceptual quality images.
GANs have been applied successfully in medical imaging~\cite{8353466,8340157,abs-1708-08333} but have not previously been used with high-resolution imaging techniques at synchrotrons. 
The challenge in the synchrotron context is that the high-resolution images produced include finely detailed features with high-frequency content. 
Approaches developed for medical images are typically insufficient since they are tailored to easily recognizable features with low-frequency content and, when applied to high-resolution images, can introduce undesired artifacts such as nonexistent features. 

Our GAN-based method, \TOMOGAN{}, adapts the U-Net network architecture\cite{8353466} to meet the specialized requirements of improving the quality of images generated by high-resolution tomography experiments at synchrotron light sources. 
We demonstrate that the \TOMOGAN{} model can be trained with limited data, performs well with high-resolution datasets, and generates greatly improved reconstructions of low-dose and noisy data, as shown in \autoref{fig:teaser-simu} and \autoref{fig:teaser-real} (with high-frequency content).
We also show that our model can be applied to a variety of experimental datasets from different instruments, showing that it is resilient to overfitting and has wider applicability in practice.

We extensively evaluate our approach with real-world tomography datasets in order to prove the applicability of the proposed method in practice. 
These experimental datasets are from different types of shale samples collected at different facilities by using the same technique but different imaging conditions (different x-ray sources and detectors).
We simulate two low-dose scenarios\cite{Du:18}: 
(1) picking a subset of the x-ray projections, to simulate reduced number of projections as in the case of a lab-based CT system,  and 
(2) applying synthetic noise to the x-ray projections, to simulate short exposure times. 
Both scenarios lead to noisy reconstructed images.
We use one dataset to train \TOMOGAN{} and then evaluate the trained model on others. 
We compare the denoised (DN) images with ground truth and measure the quality of denoised images using (1) the structural similarity (SSIM)~\cite{ssim,Ching:2017:xdesign,scikit:2014} index and (2) image pixel value plots. 
Our evaluation results show that our approach can significantly improve image quality by reducing the noise in reconstructed images.
We believe that this approach will also be effective for improving reconstruction quality when the same sample structure is imaged with different techniques with different imaging contrasts, for example, in multimodal imaging systems. 

\section{Methods}\label{sec:mdl}
We describe in turn the \TOMOGAN{} model architecture, the process by which we train a \TOMOGAN{} model, and the datasets and experimental setup used for evaluation.

\subsection{Model architecture}
Generally, the task of denoising a reconstructed image can be posed as that of translating the noisy image into a corresponding output image that represents exactly the same features, with the features in the enhanced image indistinguishable from those in a ground truth version.
Machine learning models learn to minimize a loss function---an objective that scores the quality of results---and although the learning process is automatic, the model still must be \emph{told} what needs to be minimized.

A GAN combines two neural networks, a generator ($G$) and a discriminator ($D$), which compete in a zero-sum game:  $G$ generates candidates that $D$ evaluates;  those evaluations serve as feedback to $G$. 
Thus, GANs are designed to reach a Nash equilibrium at which neither  of the two networks can reduce its costs without changing the other player's parameters.
In this paper, we train a conditional GAN in which the generator model $G$ maps a noisy reconstruction (i.e., conditionally use the noisy reconstruction as input to the $G$ instead of a random value, as it in a standard GAN~\cite{2014arXiv1406.2661G}) into a form
that is indistinguishable from the adversarial model $D$ that is trained to distinguish reconstructions of noisy projections from the enhanced noisy reconstructions created by $G$.
Thus, we use $G$ to enhance images; $D$ simply works as a helper to train the $G$. 
A classic GAN generates samples from random noise inputs~\cite{2014arXiv1406.2661G}; In contrast, our \TOMOGAN{} network creates samples from closely related noisy inputs.

\subsubsection{Generator}
\begin{figure*}[htb]
\centering
\includegraphics[width=\textwidth]{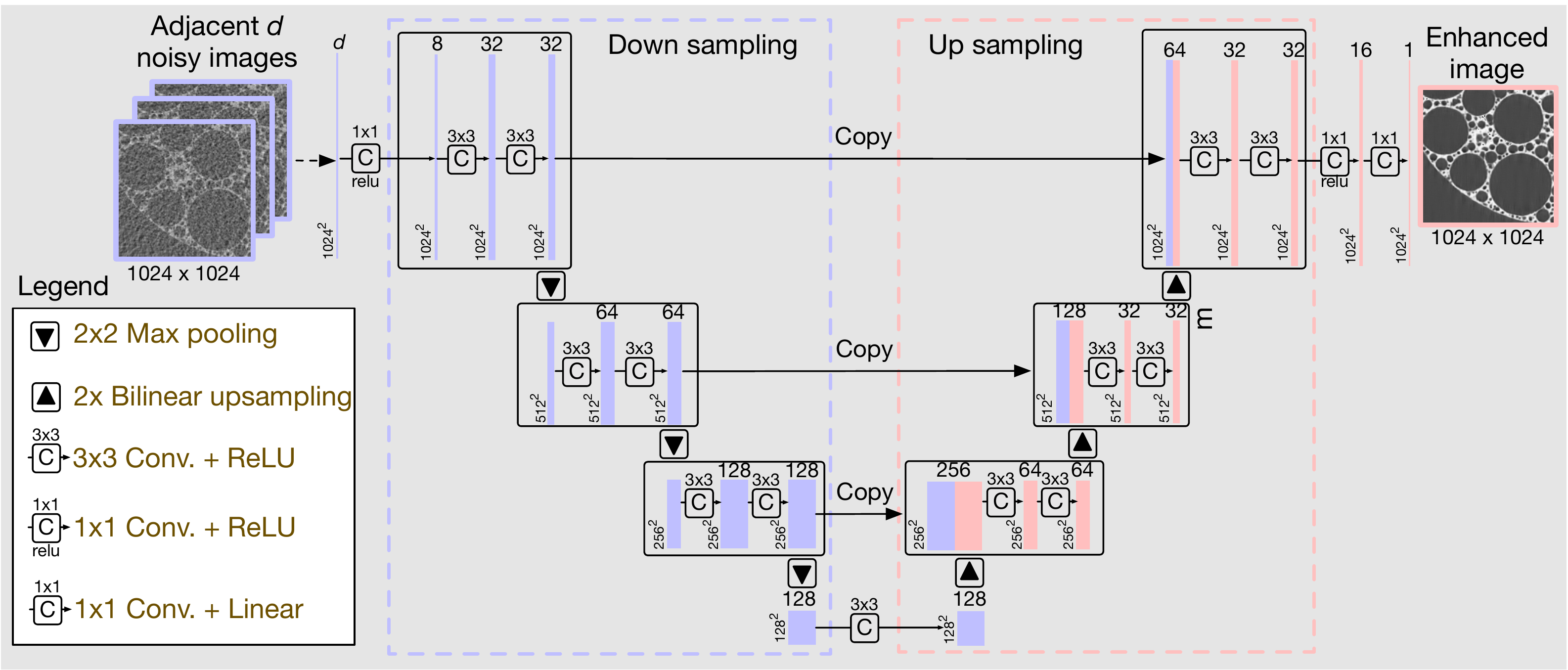}
\caption{The \TOMOGAN{} generator architecture is applied to images of size $m\times n$.
Here, we show its operation when $m=n=$1024 as an example.
Each bar corresponds to a multi-channel feature map, with the number of channels shown on the top of the bar and the height and width at the lower left edge.
The symbols (see legend) show the operations used to transform one feature map to the next.
Zero padding is used for all convolution layers to keep the output size the same as the input size. 
The \texttt{Copy} operations allow low-level information to shortcut across the network, thus improving the spatial resolution of the corresponding features.}
\label{fig:unet}
\end{figure*}

The \TOMOGAN{} generator network architecture, shown in \autoref{fig:unet}, is a variation of the U-Net architecture proposed for biomedical image segmentation by Shan et al.\cite{8353466}.
It comprises a down-sampling network (left) followed by an up-sampling network (right).
It adapts the U-Net architecture in three main ways: (1) there are three (instead of four) down-sampling layers and up-sampling layers; (2) all convolution layers have zero padding in order to keep the same image size; and (3) the input is a stack of $d$ adjacent images (discussed in the next section), and eight $1\times1$ convolution kernels are applied to the input.

In the down-sampling process, three sets of two convolution kernels (the three boxes) extract feature maps.
Then, followed by a pooling layer,  the feature map projections are distilled to the most essential elements by using a signal maximizing process.
Ultimately, the feature maps are 1/8 of the original size: 128$\times$128 in \autoref{fig:unet}.
Successful training should result in 128 channels in this feature map, retaining important features, which in \autoref{fig:activation-map} seems to be the case.

In the up-sampling process, bi-linear interpolation is used to expand feature maps. 
At each layer, high-resolution features from the down-sampling path (transmitted via \texttt{Copy} operations) are concatenated to the up-sampled output from the layer below
to form a large number of feature channels.
This structure allows the network to propagate context information to higher-resolution layers, so that the following convolution layer can learn to assemble a more precise output based on this information.

\subsubsection{Discriminator}
The \TOMOGAN{} discriminator has six 2D 3$\times$3 CNN  layers and two fully connected layers. 
Each CNN layer is followed by a leaky rectified linear unit as the activation function. 
Following the same logic as in $G$, all convolutional layers in $D$ have the same small 3$\times$3 kernel size. 
Let \texttt{CkSs-n} denote a convolution layer with a kernel size of $k\times k$, a stride of $s\times s$, $n$ output channels, and leaky ReLU activation function. 
The discriminator network consists of \texttt{C3S1-64}, \texttt{C3S2-64}, \texttt{C3S1-128}, \texttt{C3S2-128}, \texttt{C3S1-256}, \texttt{C3S2-4}, one hidden fully connected layer with 64 neurons and leaky ReLU activation, and an output layer with one neuron and linear activation. 
There is no sigmoid cross-entropy layer at the end of the discriminator, and the Wasserstein distance is used for calculating generator loss in order to improve the training stability\cite{2017arXiv170107875A}.

\subsection{Model training}\label{sec:mdl-training}
We present the loss functions used in the \TOMOGAN{} discriminator and generator.

\subsubsection{Discriminator loss}
In addition to the original critical loss presented by Arjovsky et al.\cite{2017arXiv170107875A}, we add a gradient penalty as suggested by Gulrajani et al.\cite{improved-wgan} to \autoref{eq:d-loss} for better training stability.
Thus, the discriminator is trained to minimize
\begin{equation}
\begin{aligned}
\centering
L\left( { \theta  }_{ D } \right) = &\frac { 1 }{ m } \sum _{ i=1 }^{ m }{ \left[ D\left( G\left( { I }_{ LD }^{ i } \right)  \right) -D\left( { I }_{ ND }^{ i } \right)  \right]  }  + \\
&{ \lambda  }_{ D }\frac { 1 }{ m } \sum _{ i=1 }^{ m }{ \left[ { \left( { \left\| { \nabla  }_{ \overline { I }  }D\left( { \overline { I }  }^{ i } \right)  \right\|  }_{ 2 }-1 \right)  }^{ 2 } \right],  } 
\label{eq:d-loss}
\end{aligned}
\end{equation}

\noindent
where ${ \theta  }_{ D }$ is the trainable weight/parameter of $D$, ${ I }_{ LD }^{i}$ is one noisy image ($i$th in the minibatch), $m$ is the training minibatch size, and ${ I }_{ ND }^{i}$ is the corresponding ground truth image. 

${ \overline { I }  }^{ i }=\epsilon^{i} G\left( { I }_{ LD }^{ i } \right) +\left( 1-\epsilon^{i}  \right) { I }_{ ND }^{ i }$, where $\epsilon$ is a random number between 0 and 1. 
As in Gulrajani et al.\cite{improved-wgan}, we use $\lambda_{D}=10$ to balance the trade-off between the Wasserstein distance and the gradient penalty. 

\subsubsection{Generator loss}
An important requirement while denoising images is that the generator not introduce artificial features. 
Thus, previous approaches have found it beneficial to extend the loss function to be minimized by the GAN generator with a more traditional loss, such as the ${ \ell }_{ 2 }$-norm distance (i.e., mean squared error)\cite{Nasrollahi2014}.
${ \ell }_{ 2 }$-norm distance is one of the most commonly used loss metrics for image restoration problems~\cite{sr-survey-19,Bazrafkan2019DeepLB} although it is well known to produce blurry results on image generation problems in classical approaches~\cite{Tibbs:18,Li:15}. 
Thus, we also add ${ \ell }_{ 2 }$-norm loss for the generator to enforce correctness of low-frequency structures~\cite{pix2pix2017,KimLL15a,loss-cmp}.
The discriminator's job remains unchanged, but the generator is tasked with not only 
generating adversarial samples but also being near the ground truth output in an ${ \ell }_{ 2 }$-norm sense. 
Moreover, perceptual losses are also used to penalize any structure that differs between output and target. 
Thus, the generator loss is a weighted average of three losses: the original GAN adversarial loss ${ \ell }_{ adv }$, perceptual loss ${ \ell }_{ VGG }$, and pixelwise mean ${ \ell }_{ 2 }$-norm $\ell_{mse}$.
\paragraph{Adversarial loss.}
As in the Wasserstein GAN~\cite{2017arXiv170107875A}, we compute the adversarial loss as 
\begin{ceqn}
\begin{align}
{ \ell }_{ adv }\left( { \theta  }_{ G } \right) =-\frac { 1 }{ m } \sum _{ i=1 }^{ m }{ D\left( G\left( { I }_{ LD }^{ i } \right)  \right).  } 
\end{align}
\label{eq:g-loss-adv}
\end{ceqn}

\paragraph{Perceptual loss.}
To allow the generator to retain a visually desirable feature representation, we also use the mean squared error (MSE) of features extracted by a ImageNet pre-trained VGG-19 network for the perceptual loss.  
Specifically, we use the first 16 layers of the pre-trained VGG\cite{vgg} network to extract the feature representation of a given image. 
We then define the perceptual loss as the Euclidean distance between the feature representations of a ground truth image ${ V }_{ { \theta  }_{ vgg } }\left( { I }^{ ND } \right) $ and the corresponding denoised image ${ V }_{ { \theta  }_{ vgg } }\left( { { G }_{ { \theta  }_{ G } }\left( { I }^{ LD } \right)  } \right) $. Thus, the perceptual loss is calculated by 
\begin{ceqn}
\begin{align}
{ \ell }_{ vgg }=\sum _{ i=1 }^{ { W }_{ f } }{ \sum _{ i=1 }^{ { H }_{ f } }{ { \left( { { V }_{ { \theta  }_{ vgg } }\left( { I }^{ ND } \right)  }_{ i,j }-{ V }_{ { \theta  }_{ vgg } }\left( { { G }_{ { \theta  }_{ G } }\left( { I }^{ LD } \right)  } \right) _{ i,j } \right)  }^{ 2 } } , } 
\end{align}
\label{eq:l-vgg}
\end{ceqn}

\noindent
where ${W}_{ f }$ and ${ H }_{ f }$ denote the dimensions of the feature maps extracted by the pre-trained VGG network.

Since the VGG network is trained with natural images, namely, ImageNet, one may have concerns about how well it can perform on  light source images.
Yang et al.\cite{8340157} demonstrated that the VGG network, when pre-trained with the ImageNet dataset, can also serve as a good feature extractor for CT images.

\paragraph{Pixel-wise MSE.}
The pixel-wise MSE loss is calculated as
\begin{ceqn}
\begin{align}
{ \ell }_{ mse }=\sum _{ c=1 }^{ W }{ \sum _{ r=1 }^{ H }{ { \left( { I }_{ c,r }^{ ND }-{ { G }_{ { \theta  }_{ G } }\left( { I }^{ LD } \right)  }_{ c,r } \right)  }^{ 2 } } ,  } 
\end{align}
\label{eq:l-mse}
\end{ceqn}

\noindent
where \emph{W} and \emph{H} denote the width and height of the image, respectively. 
The final loss function that the generator must minimize is thus 
\begin{ceqn}
\begin{align}
{ \ell }^{ G }={ \lambda  }_{ g }{ \ell }_{ adv }+{ \lambda  }_{ p }{ \ell }_{ mse }+{ \lambda  }_{ v }{ \ell }_{ vgg }.
\end{align}
\label{eq:l-gen}
\end{ceqn}

\begin{figure*}[hptb]
\centering
\includegraphics[width=1\textwidth]{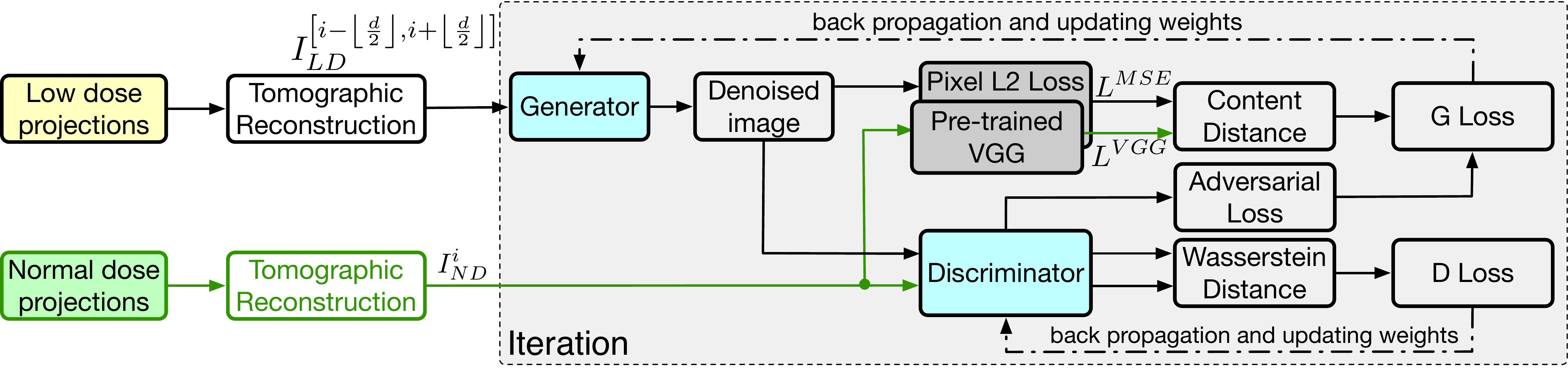}
\caption{Model training pipeline. The parameter $i$ is chosen randomly for each training iteration, and the corresponding slices of the reconstructed low-dose ($I_{LD}$) and normal-dose ($I_{ND}$) projections are used as input.
Once the model is trained, only the generator, $G$, is used to advance the tomographic reconstructions.}
\label{fig:pipeline}
\end{figure*}

To reduce the chance of ``mode collapse"\cite{2014arXiv1406.2661G}, we train $G$ once every four training steps of $D$. 
\autoref{fig:pipeline} shows the training pipeline of the model. 
Once the GAN model is trained, we can input a noisy image to the generator and it outputs the corresponding enhanced image.
In practice, we save the generator and publish it on the DLHub~\cite{dlhub18} to serve for users. 

\subsection{Datasets and experimental setup}\label{sec:exp}
We used two simulated datasets and four experimental datasets to evaluate the proposed \TOMOGAN{}. 
The experimental datasets are two shale samples that are imaged at two different facilities\cite{B19,B18}). 
The datasets are provided for benchmarking purposes and retrieved from TomoBank\cite{DeCarlo:2018:tomobank}.
 
\subsubsection{Simulated datasets}
Each simulated dataset\cite{jimaging4110128} has \num{100000} different-sized foam features distributed randomly in the 3D volume.
We use the ASTRA toolbox\cite{ASTRA} to generate the corresponding x-ray projections, with the total number of simulated projections (i.e., the total number of angles) set to \num{1024} and each projection consisting of \num{1024}$\times$\num{1024} pixels. 
We reconstruct the 3D volumes, size \num{1024}$\times$\num{1024}$\times$\num{1024}, with the filtered back projection algorithm in the TomoPy and ASTRA toolkits\cite{gursoy2014tomopy, Pelt:pp5084}, and we use the generated 2D slices in the 3D volumes for training and testing.  
Two different random seeds are used to generate two different datasets, $\emph{DS}_{simu}^{1}$ and $\emph{DS}_{simu}^{2}$.
We thus train $G$ with $\emph{DS}_{simu}^{1}$ and evaluate the trained model on $\emph{DS}_{simu}^{2}$. 

To simulate the case with a limited number of x-ray projections, we subsample projections in $\emph{DS}_{simu}^{1}$ and $\emph{DS}_{simu}^{2}$ to 1/2, 1/4, 1/8, and 1/16 of the total \num{1024} projections.
For the case with a shorter exposure time, we simulate the measured photon counts simulated by using the Beer-Lambert law~\cite{Du:18}. 
Specifically, background per-pixel photon counts are initially set to 100, 500, \num{1000}, and \num{10000}, to simulate x-ray intensity. 
Then, the photon counts are re-sampled by using Poisson distribution~\cite{Reiffen:63}, and the new noisy measurements are used to generate the line integrals and noisy image reconstructions. 
The photon absorption of the samples is set to 2.5\% for all datasets.

\subsubsection{Experimental datasets}
Shale is a challenging material because of its multi-phase composition, small grain size, low but significant amount of porosity, and strong shape- and lattice-preferred orientation. 
In this work, we use two shale sample datasets obtained from the North Sea (sample \emph{N1}\/) and the Upper Barnett Formation in Texas (sample \emph{B1}\/)\cite{B18}.
Each sample has been imaged at two different light source facilities: the Advanced Photon Source (APS) at Argonne National Laboratory and the Swiss Light Source (SLS) at the Paul Scherrer Institut.
For ease of reference, we name the resulting four datasets as $\emph{DS}_{APS}^{B1}$, $\emph{DS}_{APS}^{N1}$, $\emph{DS}_{SLS}^{B1}$, and $\emph{DS}_{SLS}^{N1}$. 

When evaluating performance with a limited number of projections, we subsample to 1/2, 1/4, 1/8, and 1/16 of the total number of projections in each full dataset.
Similar to simulated datasets, we apply Poisson noise \cite{Reiffen:63,boas2012ct} to simulate the impact of shorter exposure times on projection quality, and we add this simulated noise to the normal exposure time projections. 

We arbitrarily select the dataset $\emph{DS}_{APS}^{B1}$ to train \TOMOGAN{} and use the other three datasets to evaluate its effectiveness and performance. 
To train or enhance the $i^{th}$ slice, we use slices from $i-\left\lfloor d/2 \right\rfloor$ to $i+\left\lfloor d/2 \right\rfloor$ as input to the generator $G$, where $d$ is a tunable parameter. 
The generated (enhanced) images as well as the corresponding $i^{th}$ slice of the normal dose are then input to the discriminator to compute the loss and update $\theta_D$.
We implemented \footnote{Our open source implementation is available at \url{git@github.com:ramsesproject/TomoGAN.git}} \TOMOGAN{} with Tensorflow\cite{abadi2016tensorflow} and used one NVIDIA Tesla V100 GPU cards for training.
We used the ADAM algorithm\cite{Adam} to train both the generator and discriminator, with a batch size of four images and a depth of three for each image.

\section{Experimental Results}\label{sec:exp-res}
In this section, we evaluate the performance of  \TOMOGAN{} with different configurations and various datasets. 
We first show the effectiveness of the depth parameter in \TOMOGAN{} and importance of the various loss terms. 
Next, we present the performance of \TOMOGAN{} with both simulation and experimental datasets. 
We then compare the computational requirements and image quality of FBP followed by \TOMOGAN{} against a simultaneous iterative reconstruction technique.
At the end, we compare \TOMOGAN{} with other representative solutions on experimental dataset.

\subsection{Effectiveness of using adjacent slices in image enhancement}\label{sec:eff-depth}
\begin{figure*}[htbp]
\centering
\includegraphics[width=\textwidth]{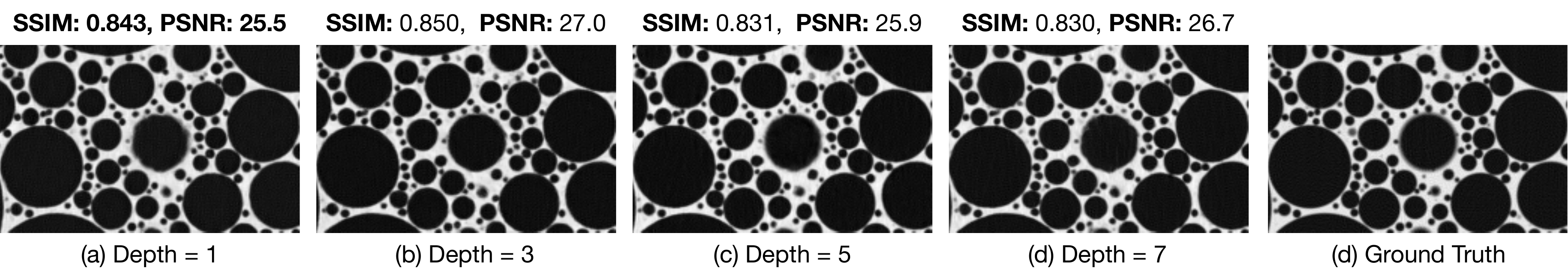}
\caption{Effectiveness of using varying numbers of adjacent slices when enhancing reconstructed images. } 
\label{fig:depth-eva-simu} 
\end{figure*}

\begin{figure}[htbp]
\centering
\includegraphics[width=0.8\textwidth]{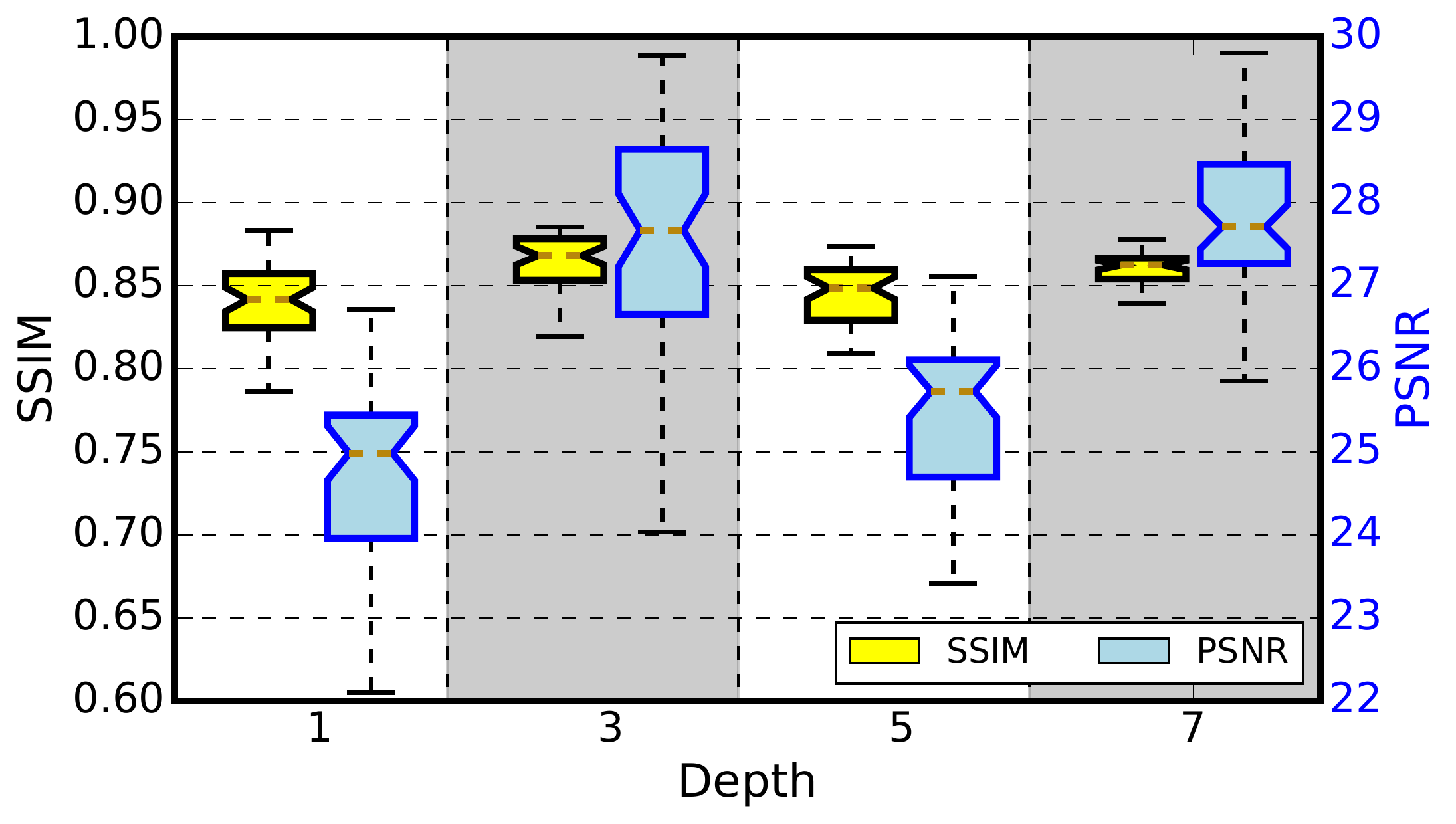}
\caption{Effectiveness of using varying numbers of adjacent slices when enhancing reconstructed images. } 
\label{fig:depth-eva-simi} 
\end{figure}

Features observed in adjacent slices tend to be highly correlated, usually with similar shape but different size. 
Stacking several neighboring slices to form multiple channels for the first CNN layer is equivalent to using a multiple linear regressor as a filter to reduce noise. 
Thus we use slices from $i-\left\lfloor d/2 \right\rfloor$ to $i+\left\lfloor d/2 \right\rfloor$ as input to enhance the $i$th slice,  where $d$ is a hyperparameter. 
To explore the effectiveness of $d$, we used $\emph{DS}_{simu}^{1}$ to train models with different $d$ (i.e., $d=1$, $d=3$, $d=5$, and $d=7$) and test their performance on $\emph{DS}_{simu}^{2}$.
\autoref{fig:depth-eva-simu}, which compares the quality of an arbitrarily chosen image enhanced by model with different $d$, shows the effectiveness of using adjacent slices. 
For all of the slices in  $\emph{DS}_{simu}^{2}$, \autoref{fig:depth-eva-simi}  compares their SSIM and PSNR.
One can see from \autoref{fig:depth-eva-simi} that $d$ has big influence on mode performance, and that $d=3$ gets the best quality, especially when the original feature edge is not sharp (e.g., the center circle in \autoref{fig:depth-eva-simu}). 
Using adjacent slices is one of the difference between our method and WGAN-VGG~\cite{8340157}.
We note that the best depth $d$ depends on dataset characteristics such as feature resolution. 
$d=3$ may not be the best for other datasets where feature sizes change slowly across slices.

\subsection{Importance of the various loss terms}\label{sec:gan-vs-sv}
\begin{figure*}[htbp]
\centering
\includegraphics[width=\textwidth]{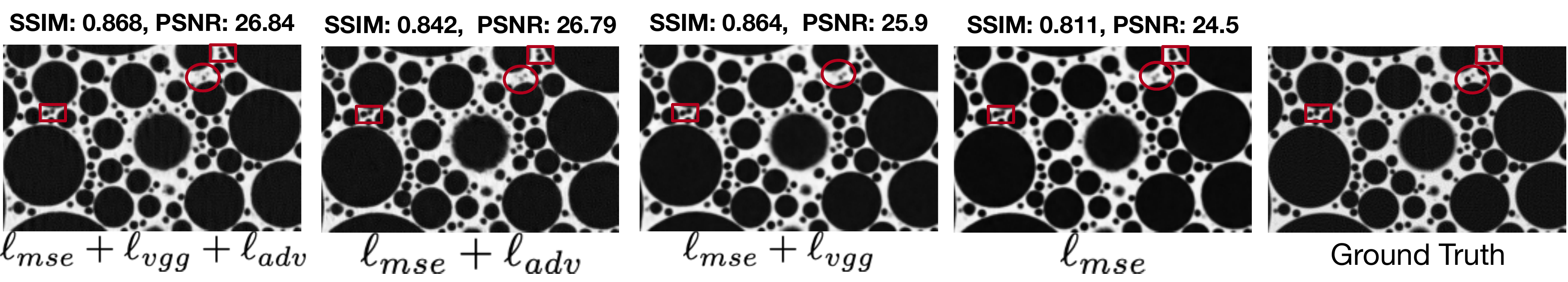}
\caption{Effectiveness of adversarial and perceptual loss.
From left to right with ${\ell}_{2}$-norm, reconstruction 
with both adversarial and perceptual loss;
with just adversarial loss; 
with just perceptual loss; 
and with neither adversarial nor perceptual loss.
} 
\label{fig:loss-eva-img} 
\end{figure*}

Previous works have used supervised machine learning techniques\cite{Yang2018,jimaging4110128,fbpconv} to learn the mapping between reconstructions from noisy and noise-free images. 
Our approach of training a GAN model and using its generator to improve reconstructions is intended to make \TOMOGAN{} more resilient to overfitting and the creation of artifacts. 
In order to evaluate how sensitive results are to the chosen loss terms, we conducted experiments that selectively disable ${\ell}_{adv}$ and ${ \ell }_{ vgg }$.
\autoref{fig:loss-eva-img} shows the results achieved on a representative slice of a \num{1024}-slice dataset.
Notice how, in the three images to the left, some small features (e.g., those highlighted by the red rectangle) are merged and others (e.g., those highlighted by the red ellipse) are difficult to see.
Such artifacts are not tolerable in practice.
${\ell}_{adv}$ and ${ \ell }_{ vgg }$ help to avoid such artifacts.

\begin{figure}[htbp]
\centering
\includegraphics[width=0.8\textwidth]{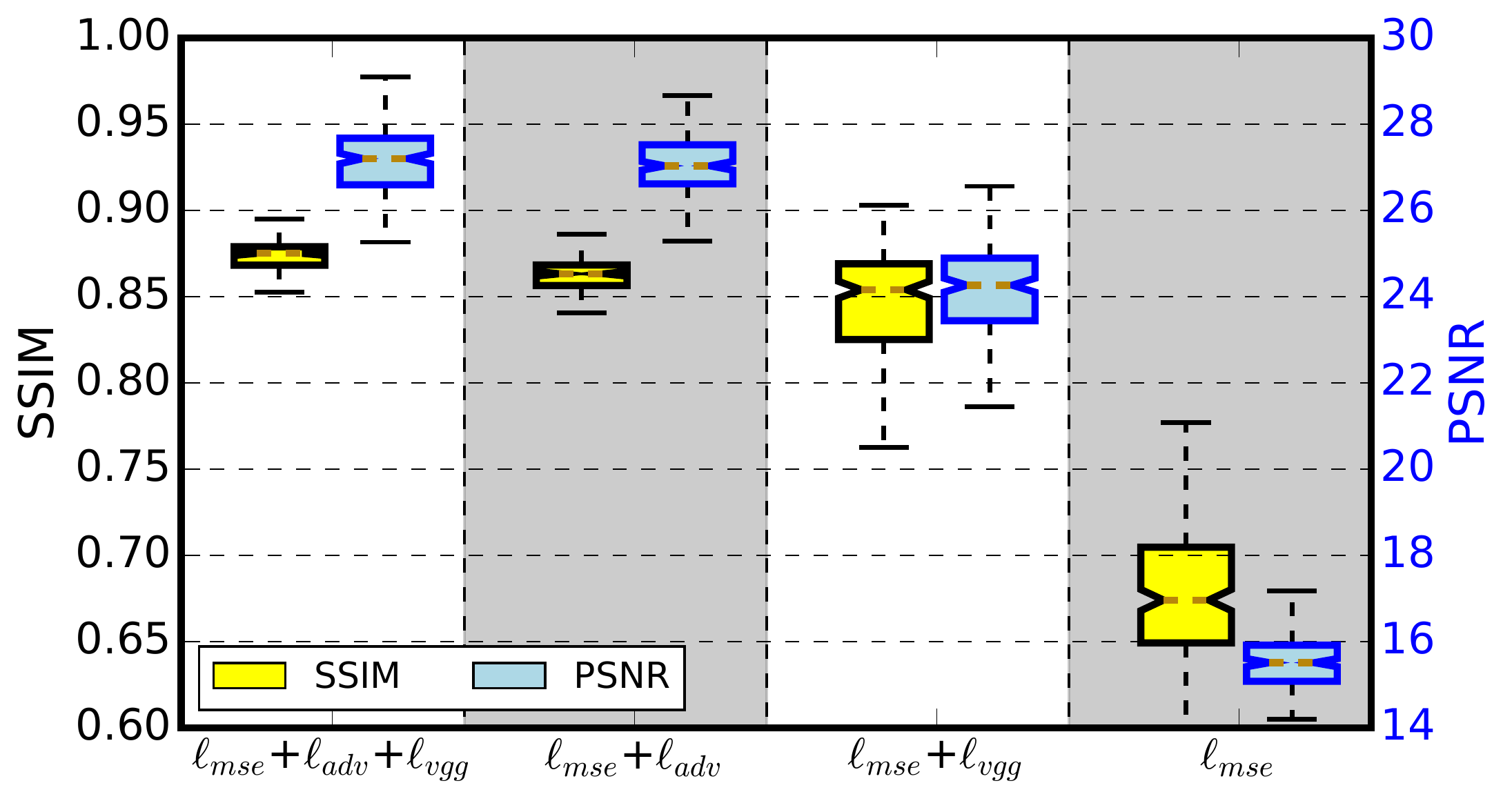}
\caption{Effectiveness of adversarial loss and perceptual loss: SSIM and PSNR statistics. } 
\label{fig:loss-eva-simi} 
\end{figure}

\autoref{fig:loss-eva-simi} provides another perspective on the effectiveness of adversarial and perceptual loss,
showing the SSIM and PSNR statistics for all slices of the same dataset.
We see that the adversarial and perceptual loss terms each provide
considerable improvements when used in isolation. 
The two together are only slightly better than adversarial loss alone.

\subsection{Simulated datasets}\label{sec:exp-res-simu}
Here, we show the effectiveness of \TOMOGAN{} in improving noisy reconstructed images with simulated datasets. 
We show both fewer projections and shorter exposure time cases that result in noisy measurement data and reconstructed images. 

\subsubsection{Fewer projections}\label{sec:simu-ss}
\begin{figure*}[ht]
\centering
\includegraphics[width=\textwidth]{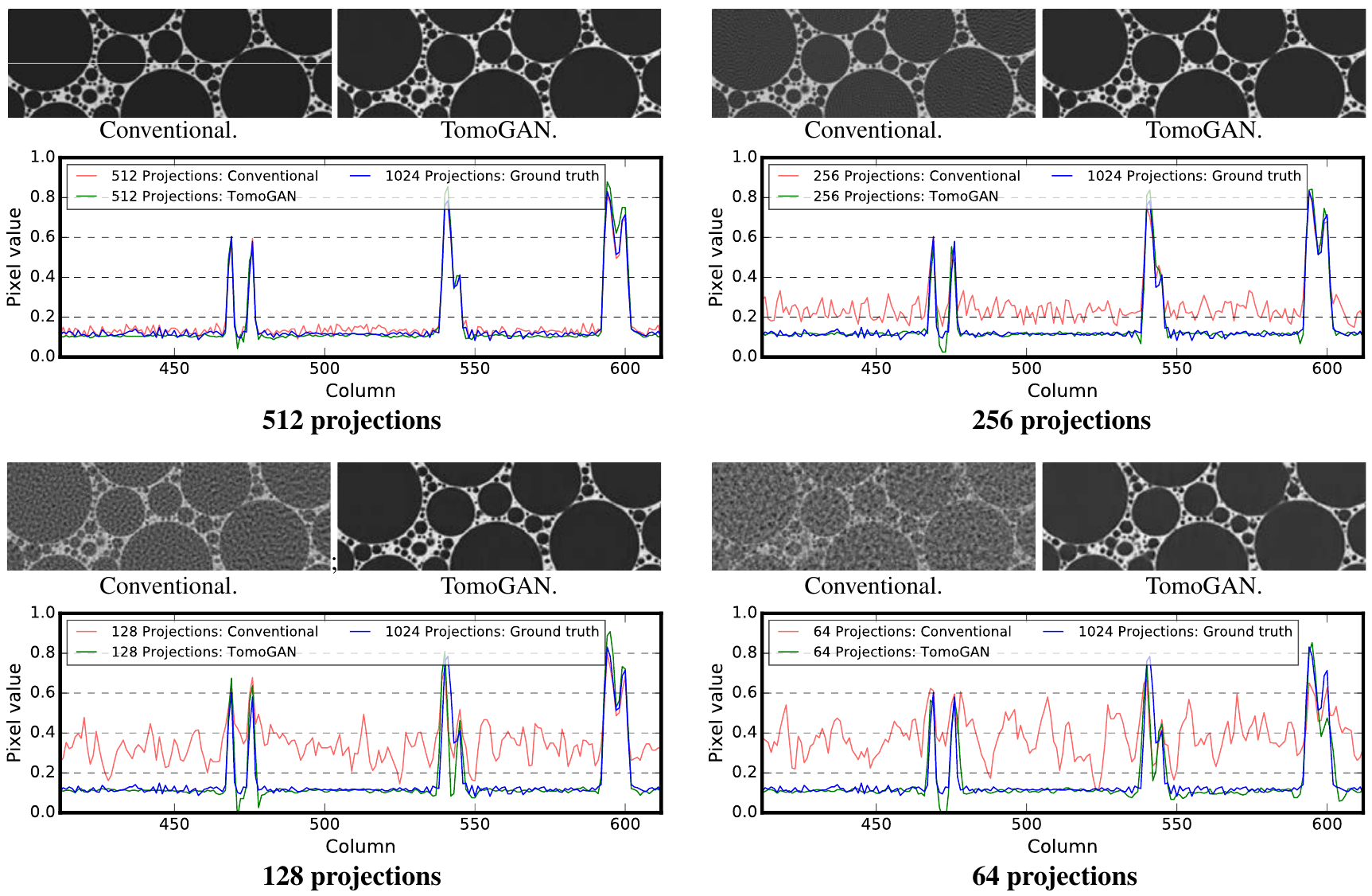}
\caption{Conventional vs.\ \TOMOGAN{}-enhanced reconstructions of simulated data, subsampled to (512, 256, 128, 64) projections. In each group of three elements, the two images show conventional and \TOMOGAN{} reconstructions, while the plot shows conventional, \TOMOGAN{}, and ground truth values for the 200 pixels on the horizontal line in the top left image.} 
\label{fig:simu-ss-ds1-combo}
\end{figure*}

In these experiments, we subsample the \num{1024} projections to 512, 256, 128, and 64 projections to simulate a scenario for the step scan CT systems where a limited number of projections are collected to reduce either  the total experiment time (e.g., to capture dynamic features) or the x-ray dose (e.g., for x-ray dose sensitive sample). 
We perform the same subsampling for both $DS_{simu}^{1}$ and $DS_{simu}^{2}$.
Then, once \TOMOGAN{} is trained with $DS_{simu}^{1}$, 
we use its generator to enhance reconstructions based on subsampled projections of $DS_{simu}^{2}$. 
In \autoref{fig:simu-ss-ds1-combo}, we zoom in on a randomly selected area and plot the pixel values of a single horizontal line  to show the improvement in image quality.  
We see that the reconstruction quality with reduced numbers of projections is poor in the absence of enhancement; however, the  enhancement with \TOMOGAN{} yields images with quality comparable to that of the reconstructions based on \num{1024} projections.

We use SSIM~\cite{ssim} to perform  quantitative comparisons of the quality of \TOMOGAN{}-enhanced reconstructions. 
Using a reconstruction of a full \num{1024}-projection dataset as ground truth, we calculate for each noisy (fewer projections or fewer photons) dataset both (a) SSIM between ground truth and the subsampled reconstruction, \emph{SSIM}$_{GT}^{LD}$, 
and (b) SSIM between ground truth and the \TOMOGAN{}-enhanced version of that subsampled reconstruction, \emph{SSIM}$_{GT}^{DN}$.
The difference between these two quantities is the image quality improvement that results from the use of \TOMOGAN{}.

We further calculated the two values \emph{SSIM}$_{GT}^{LD}$ and \emph{SSIM}$_{GT}^{DN}$ for each slice in the reconstructed image that has features, thus allowing us to plot in \autoref{fig:simu-ssim} the distribution of SSIM image quality scores for both the conventional (\emph{SSIM}$_{GT}^{LD}$) and \TOMOGAN{}-enhanced (\emph{SSIM}$_{GT}^{DN}$) images.
We see that while image quality declines significantly with subsampling in the absence of \TOMOGAN{} enhancement, it remains high when enhancement is applied, even with high degrees of subsampling.  
SSIM is improved significantly by \TOMOGAN{} and is improved consistently across all the slices, suggesting that the method is reliable.
The evaluation of this low-dose scenario can also be used for streaming tomography where the views are very limited at the beginning of experimentation (e.g., Liu et al. 2019~\cite{liu2019deep}).

\subsubsection{Shorter exposure times}\label{sec:simu-exp}
Another way of reducing dose (or, equivalently, accelerating experiments) is to reduce x-ray exposure times.
The use of shorter exposure times has attracted major attention for synchrotron-based tomography systems (e.g., Advanced Photon Source), 
because it can reduce data collection times significantly, 
as needed both to capture dynamical features and to reduce damage to organic samples from light source radiation.  
Reduced exposure time, however, compromises the signal-to-noise ratio, which affects final reconstruction image quality and may affect scientific findings.

\begin{figure*}[ht]
\centering
\includegraphics[width=\textwidth]{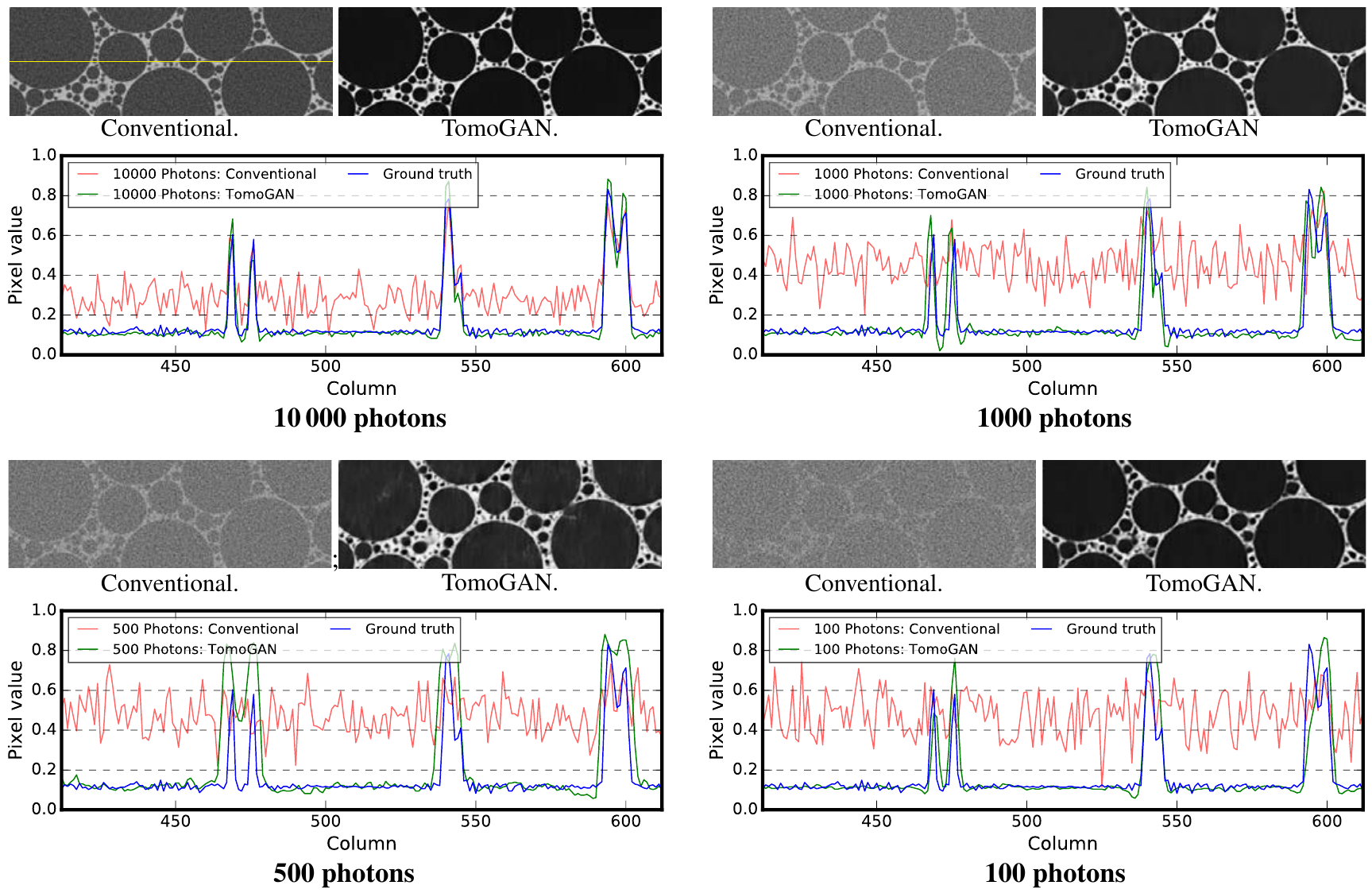}
\caption{Conventional vs.\ \TOMOGAN{}-enhanced reconstructions 
of simulated data with intensity limited to \{\num{10000}, \num{1000}, 500, 100\} photons per pixel. Figure elements are as in \autoref{fig:simu-ss-ds1-combo}.}
\label{fig:simu-exp-ds1-combo}
\end{figure*}

\begin{figure}[ht]
\centering
\begin{subfigure}{.48\linewidth}
\centering
\includegraphics[width=\textwidth,trim=0.1in 0 0 0,clip]{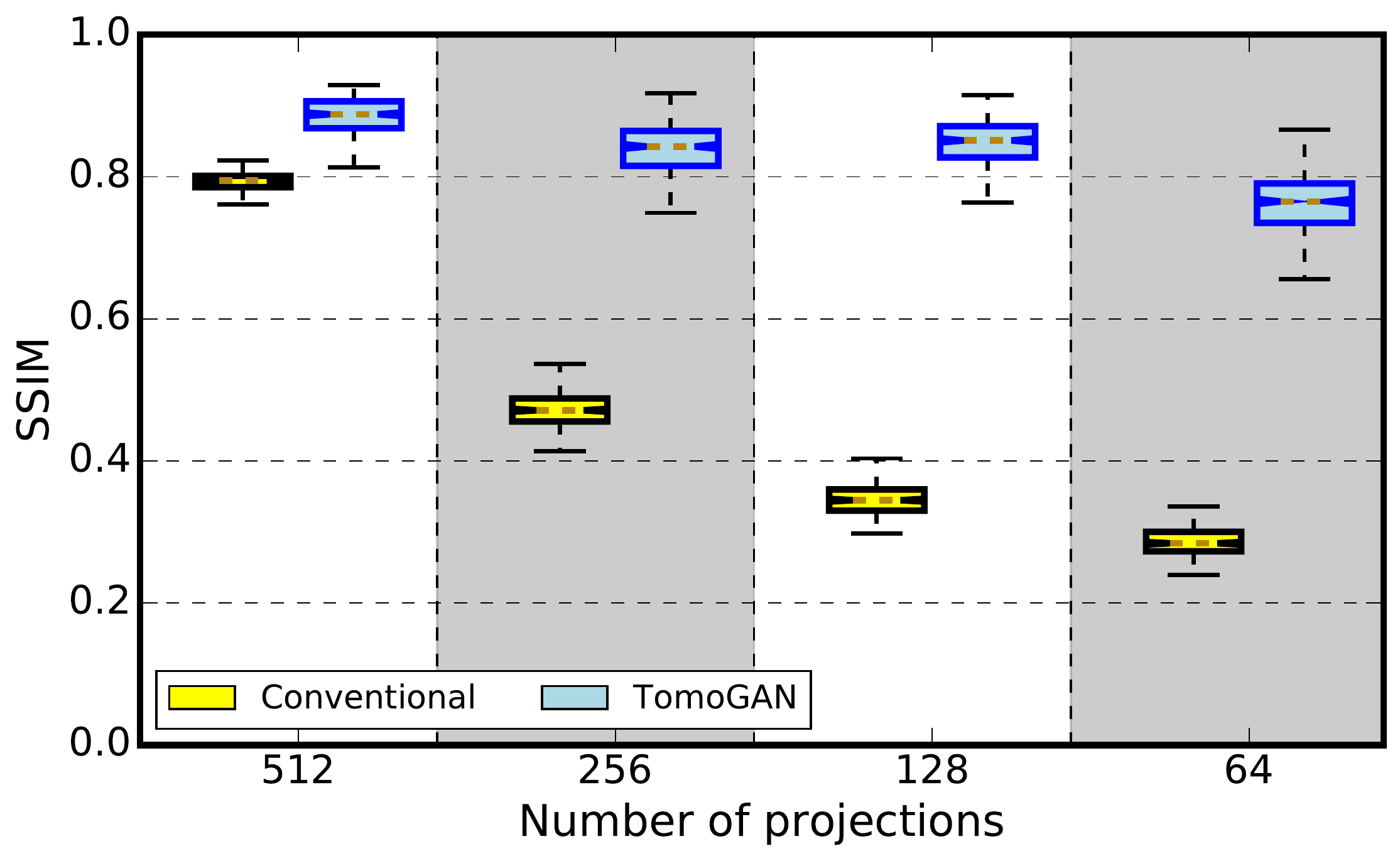}
\vspace{-4ex}
\caption{Fewer x-ray projections.}
\label{fig2:simu-ssim-ss}
\end{subfigure}
\begin{subfigure}{.48\linewidth}
\centering
\includegraphics[width=\textwidth,trim=0.1in 0 0 0,clip]{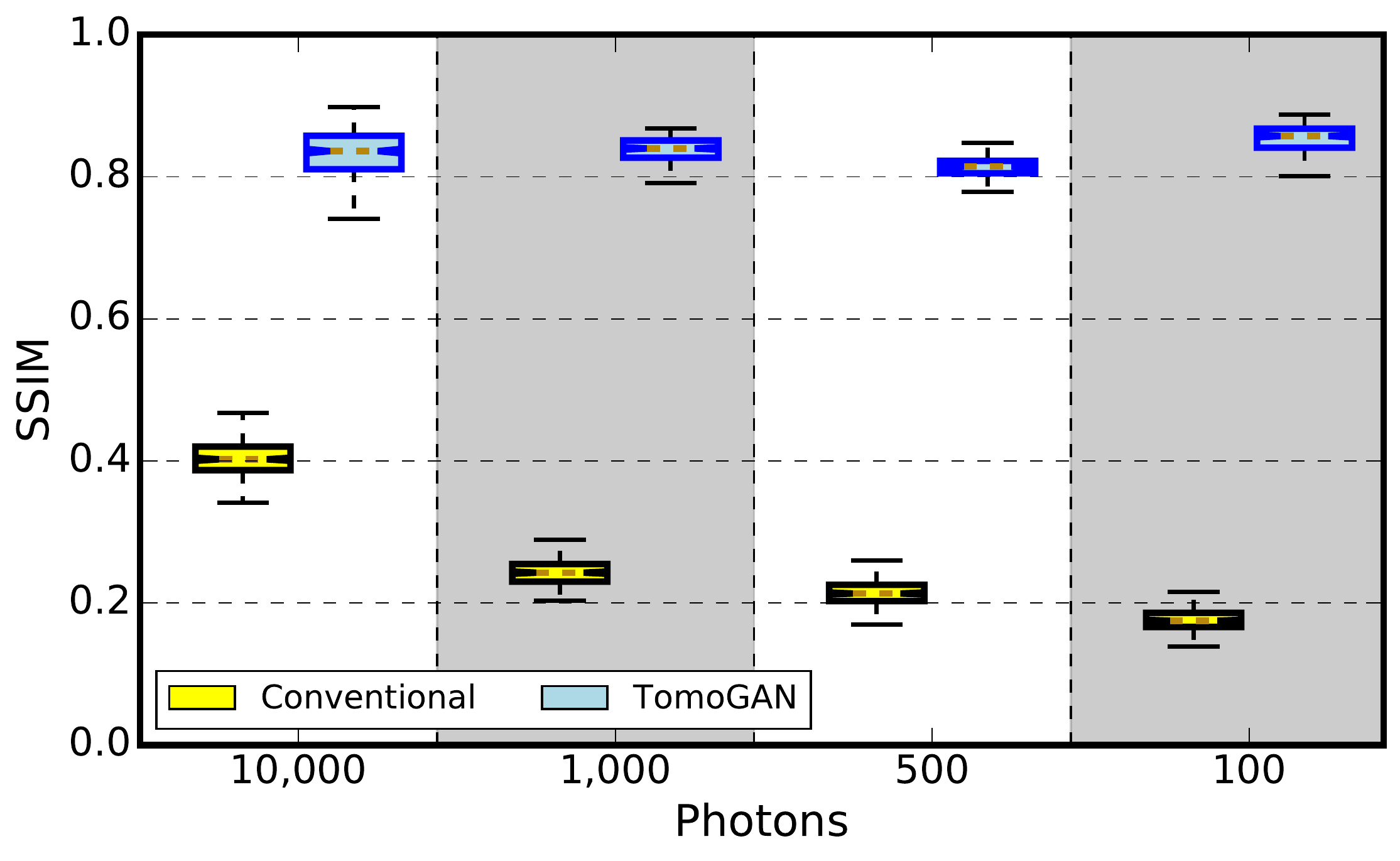}
\vspace{-4ex}
\caption{Shorter exposure time.}
\label{fig2:simu-ssim-exp}
\end{subfigure}
\caption{Per-slice SSIM similarities with ground truth simulated data,
for both conventional and \TOMOGAN{}-enhanced reconstructions and for
different degrees of subsampling. }
\label{fig:simu-ssim}
\end{figure}

In this experiment, we simulate x-ray projections with different photon intensities 
to simulate the effect of different exposure times. 
Specifically, we simulate with intensities of 100, 500, \num{1000}, and \num{10000} photons per pixel and, for each, compute conventional and \TOMOGAN{}-enhanced reconstructions.
\autoref{fig:simu-exp-ds1-combo} shows representative results, while \autoref{fig2:simu-ssim-exp} quantifies the quality of the different reconstructed images via SSIM scores computed relative to ground truth.
Ground truth here is a reconstruction based on the noise-free simulation data.
We see that the trained \TOMOGAN{} model can enhance the reconstructions of noisy data to a quality comparable with that of reconstructions based on ground truth. 
Comparison of \autoref{fig:simu-exp-ds1-combo} with \autoref{fig:simu-ss-ds1-combo} shows that reducing exposure time leads to a different kind of noise from that when using fewer 
projections.

\subsection{Experimental datasets}\label{sec:exp-res-real}
A key issue to address when dealing with experimental datasets is whether we can train \TOMOGAN{} on data collected from one sample and then use the  trained model on other samples and/or at other facilities. 
Thus, we train \TOMOGAN{} with data collected on one sample at one facility, $DS_{APS}^{\emph{B1}}$, and then evaluate its performance on noisy versions (both fewer projections and shorter exposure times) of three other datasets, $DS_{APS}^{\emph{N1}}$, $DS_{SLS}^{\emph{B1}}$, and $DS_{SLS}^{\emph{N1}}$. 
Then, we use the trained model to enhance noisy images of the other three datasets.
Since these other datasets include  different samples ($DS_{*}^{\emph{N1}}$ and $DS_{*}^{\emph{B1}}$) and x-ray projections collected at different facilities ($DS_{APS}^{\emph{*}}$ and $DS_{SLS}^{\emph{*}}$), this configuration mimics a practical use case. 

Evaluation of \TOMOGAN{} is more difficult for the experimental datasets since there is no ground truth.
Therefore, we considered the images that are reconstructed using full datasets (\num{1024}-projection) as their corresponding ground truths.

\subsubsection{Fewer projections}\label{sec:real-ss}
\begin{figure*}[ht]
\centering
\includegraphics[width=\textwidth]{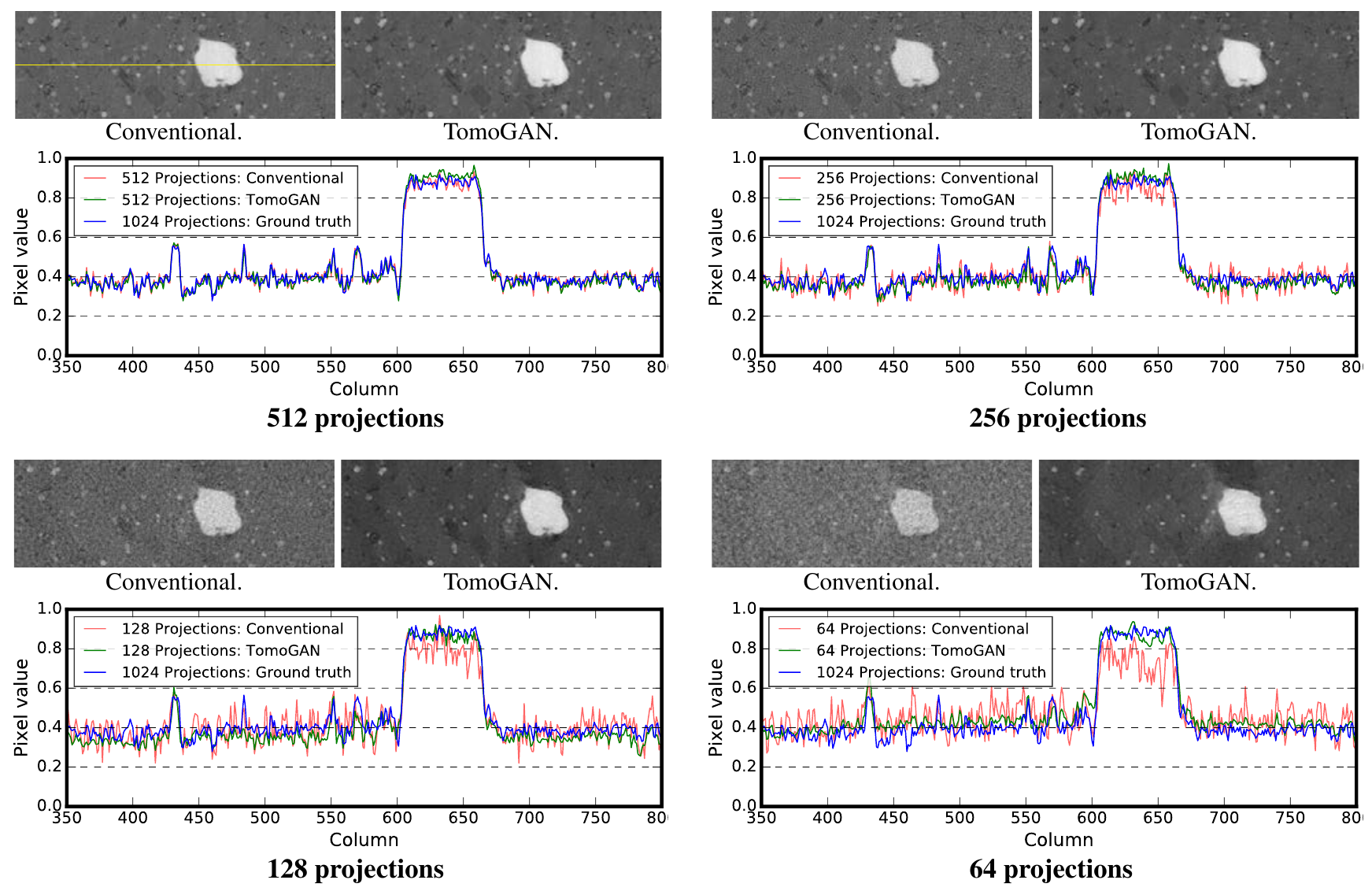}
\caption{Conventional vs.\ \TOMOGAN{}-enhanced reconstructions 
of experimental dataset $DS_{SLS}^{\emph{B1}}$, subsampled to  (512, 256, 128, 64) projections.
Figure elements are as in \autoref{fig:simu-ss-ds1-combo}.}
\label{fig:aps-ss-ds3-combo}
\end{figure*}

As we did for the simulated datasets, we subsample the full-resolution (\num{1024}-projection) experimental datasets and then use \TOMOGAN{} to enhance the reconstructions of their subsampled datasets.
In \autoref{fig:aps-ss-ds3-combo}, we show conventional and \TOMOGAN{}-enhanced reconstructions (for varying degrees of subsampling) of a randomly selected area of a slice from $DS_{SLS}^{\emph{B1}}$. The corresponding pixels of a randomly selected horizontal line within the selected areas are also presented for comparison. 
We see that the image qualities are improved significantly: even the small features are clearly visible in the enhanced images.
The pixel value plots also show that the \TOMOGAN{}-enhanced images are comparable with the ground truth.

\begin{figure}[ht]
\centering
\begin{subfigure}{0.48\linewidth}
\centering
\includegraphics[width=\textwidth]{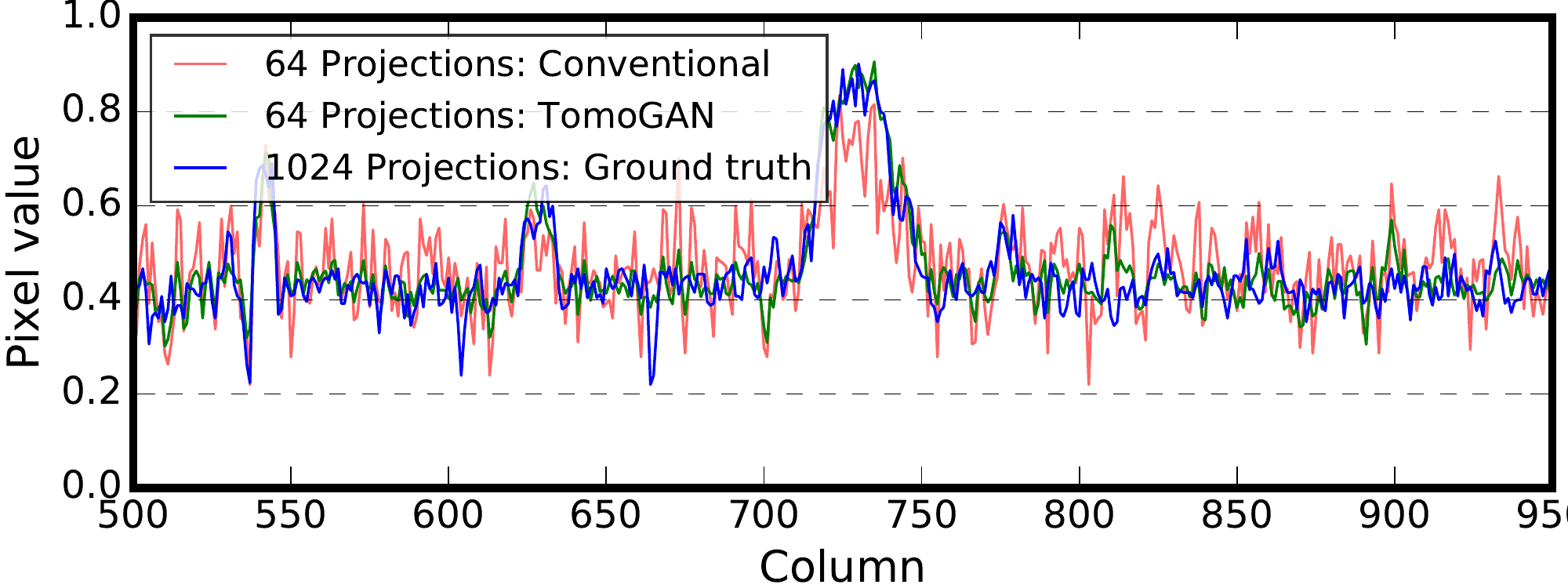}
\caption*{$DS_{APS}^{\emph{B1}}$ }
\end{subfigure}
\begin{subfigure}{0.48\linewidth}
\centering
\includegraphics[width=\textwidth]{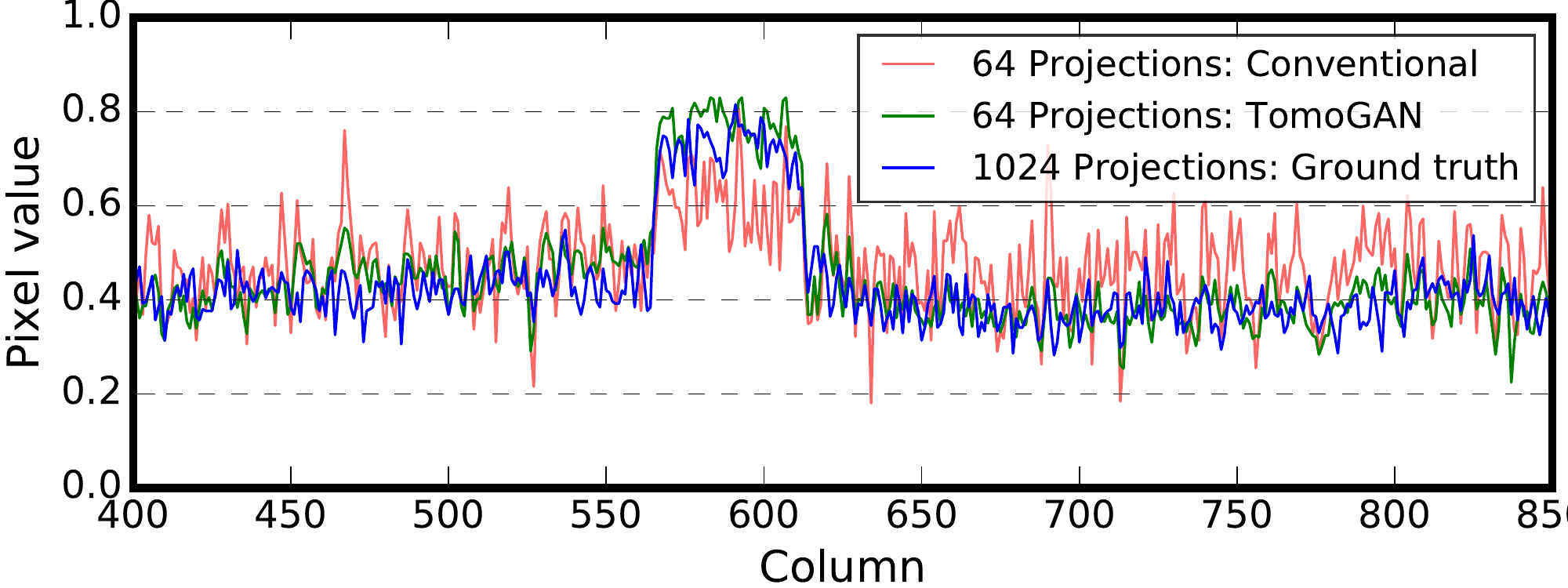}
\caption*{$DS_{\emph{APS}}^{\emph{N1}}$}
\end{subfigure}
\begin{subfigure}{0.48\linewidth}
\centering
\includegraphics[width=\textwidth]{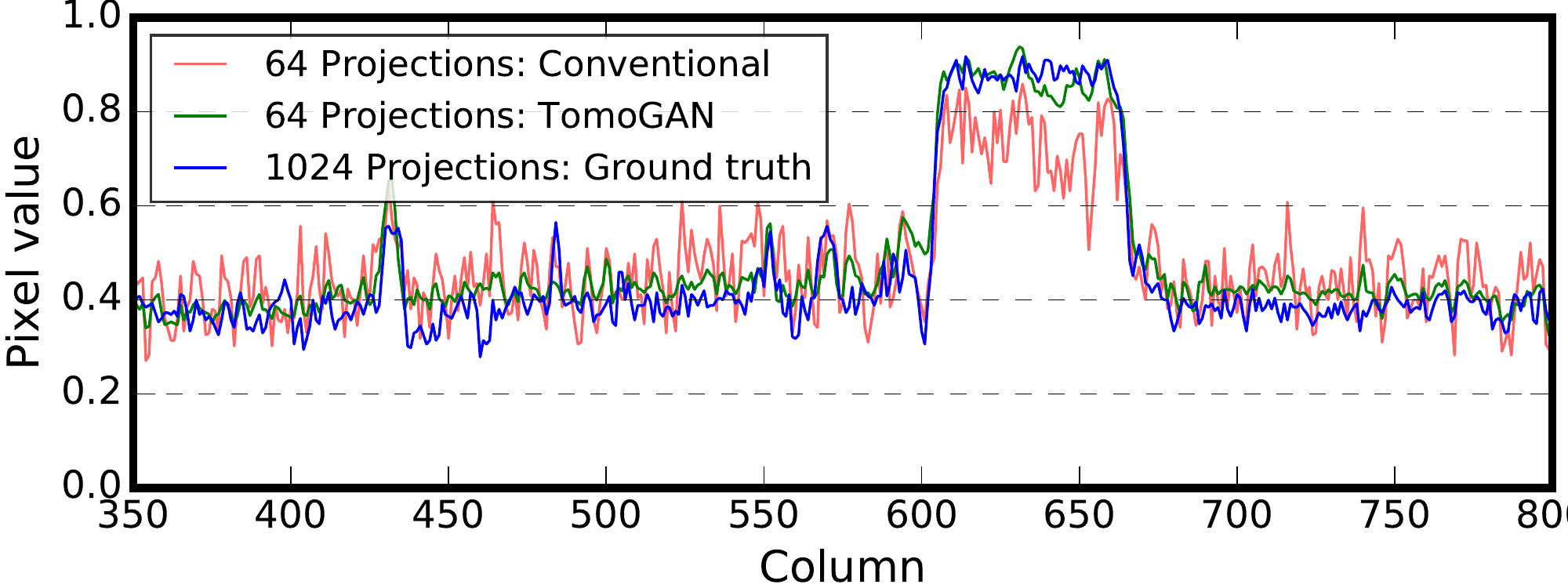}
\caption*{$DS_{SLS}^{\emph{B1}}$}
\end{subfigure}
\begin{subfigure}{0.48\linewidth}
\centering
\includegraphics[width=\textwidth]{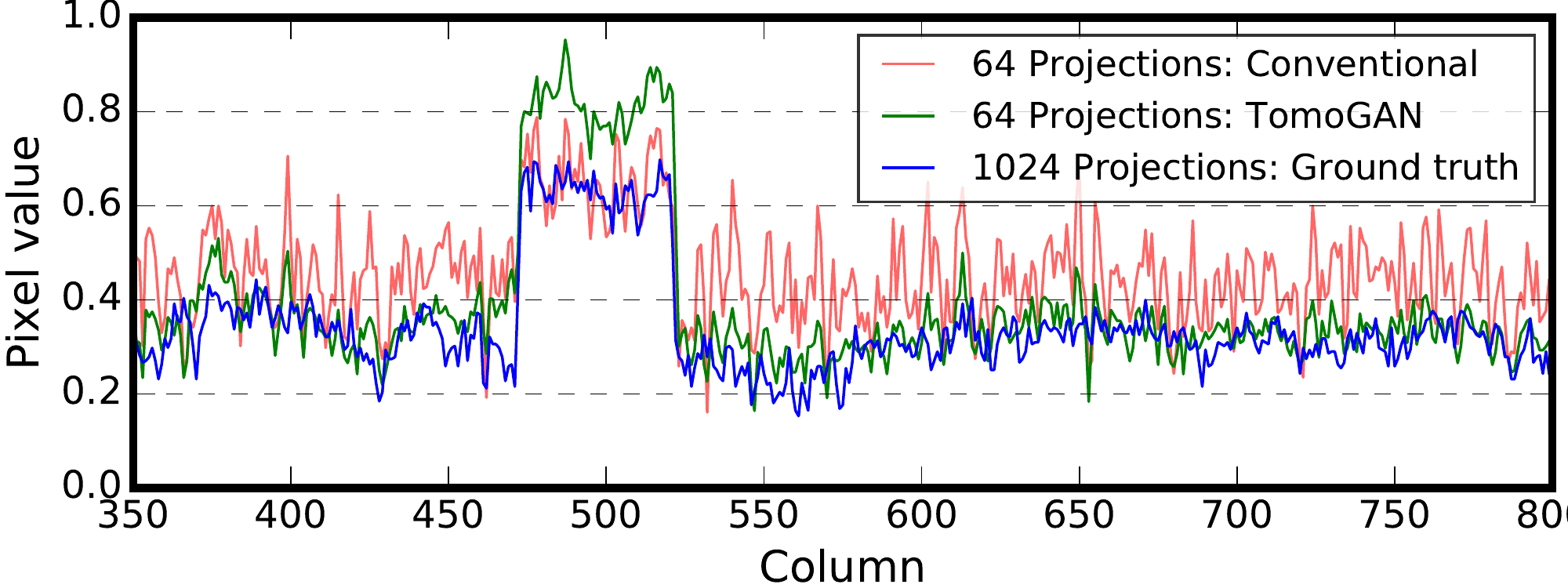}
\caption*{$DS_{SLS}^{\emph{N1}}$}
\end{subfigure}
\caption{Pixel values of an arbitrarily chosen line in each of the four experimental datasets, subsampled to 64 projections.} 
\label{fig:aps-all-ds-ss16-pixel} 
\end{figure}

For the most challenging cases, in which only 64 x-ray projections are available for reconstruction, 
\autoref{fig:aps-all-ds-ss16-pixel} shows the pixel values of a horizontal line that crosses an arbitrarily chosen feature in each of the four datasets. 
Comparing the enhanced (green) and the ground truth (blue) lines in each case, we see that \TOMOGAN{} yields reconstructions comparable to the ground truth.
The contrast of the enhanced images is clearly higher than that of the noisy reconstructions; 
and the variance of the feature pixel values (corresponding to the noise level) has been reduced considerably from the noisy reconstruction cases. 
Moreover, not only is the experiment time reduced when using fewer projections (by a factor corresponding to the subsampling ratio), but the dataset size and computation cost for reconstructions are also reduced by the same fraction. 

We again use SSIM to quantify the qualities of the conventional and \TOMOGAN{}-enhanced reconstructions.
Recall that our model was trained with $DS_{APS}^{\emph{B1}}$. 
Here we evaluate the trained model on a different shale sample imaged at a different facility: $DS_{SLS}^{\emph{N1}}$.
The SSIM metric scores, shown in \autoref{fig2:aps-ssim-ss}, indicate that \TOMOGAN{} consistently improves image quality for all slices at each subsampling level. 
However, the overall quality scores for the \TOMOGAN{}-enhanced reconstructions are clearly not as good as those for the simulated data in \autoref{fig2:simu-ssim-ss}.
We attribute this difference to the fact that the simulated dataset has a much better training dataset (ground truth) than does the experimental dataset and that the features in the simulation dataset are much simpler (only circles) than the features in the experimental dataset.

The results in \autoref{fig:aps-ssim} show metric scores that are consistent across slices, suggesting that our model behaves well for all slices.
We claim that this property is important for a black-box predictor.

\subsubsection{Shorter exposure time}\label{sec:real-exp}
We also trained \TOMOGAN{} to enhance the quality of images reconstructed from short-exposure-time projections.
First we applied Poisson noise~\cite{boas2012ct} to the four experimental datasets to create new (simulated) short-exposure-time datasets.
Next, we used one of the experimental datasets, $DS_{APS}^{\emph{B1}}$, plus its associated short exposure datasets, to train \TOMOGAN{}.
Then, we used the trained model to enhance the noisy images of the other three datasets.

\begin{figure}[ht]
\centering
\begin{subfigure}{.495\linewidth}
\centering
\includegraphics[width=\textwidth,trim=0.1in 0 0 0,clip]{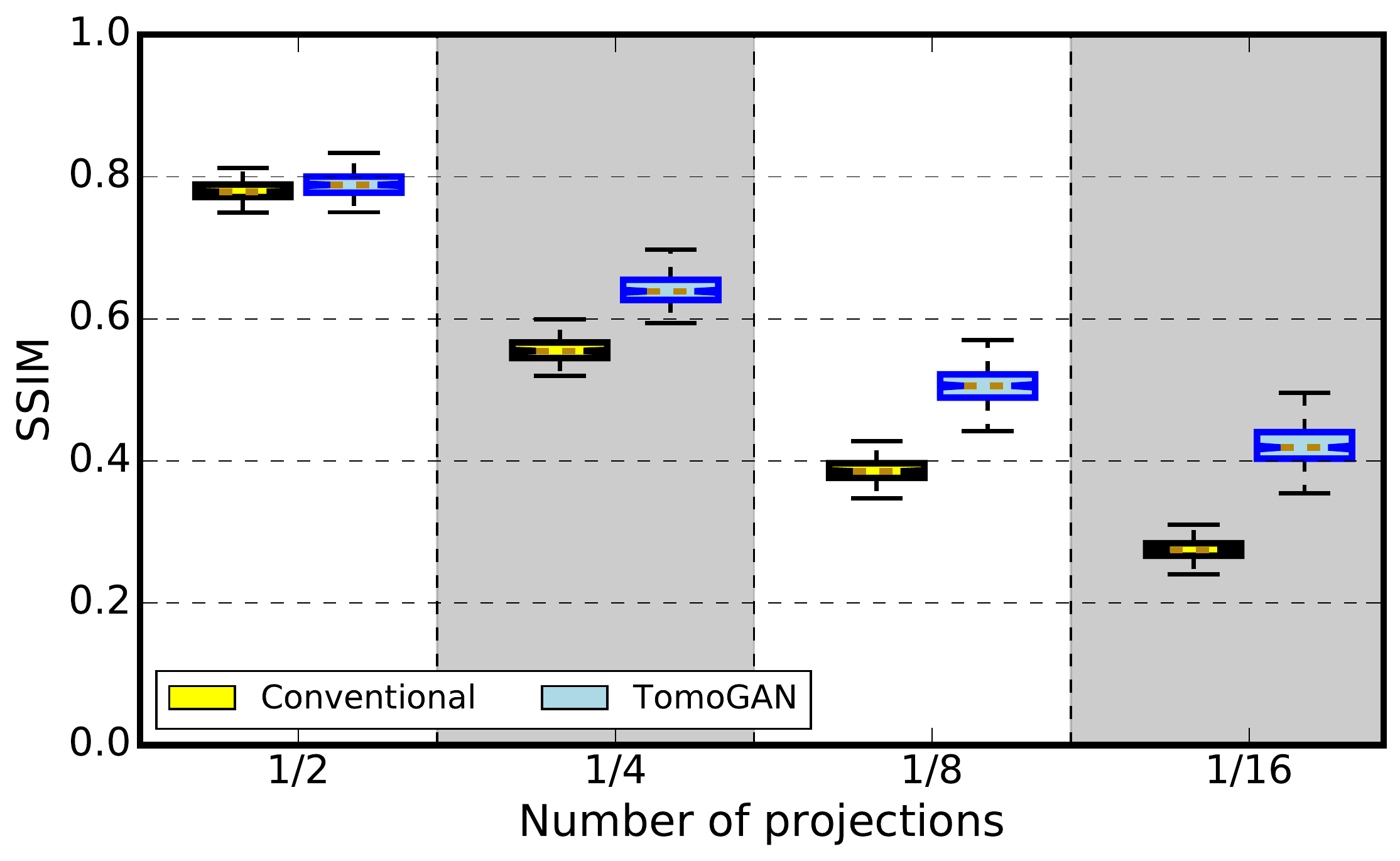}
\vspace{-4ex}
\caption{Fewer x-ray projections.}
\label{fig2:aps-ssim-ss}
\end{subfigure}
\begin{subfigure}{.495\linewidth}
\centering
\includegraphics[width=\textwidth,trim=0.1in 0 0 0,clip]{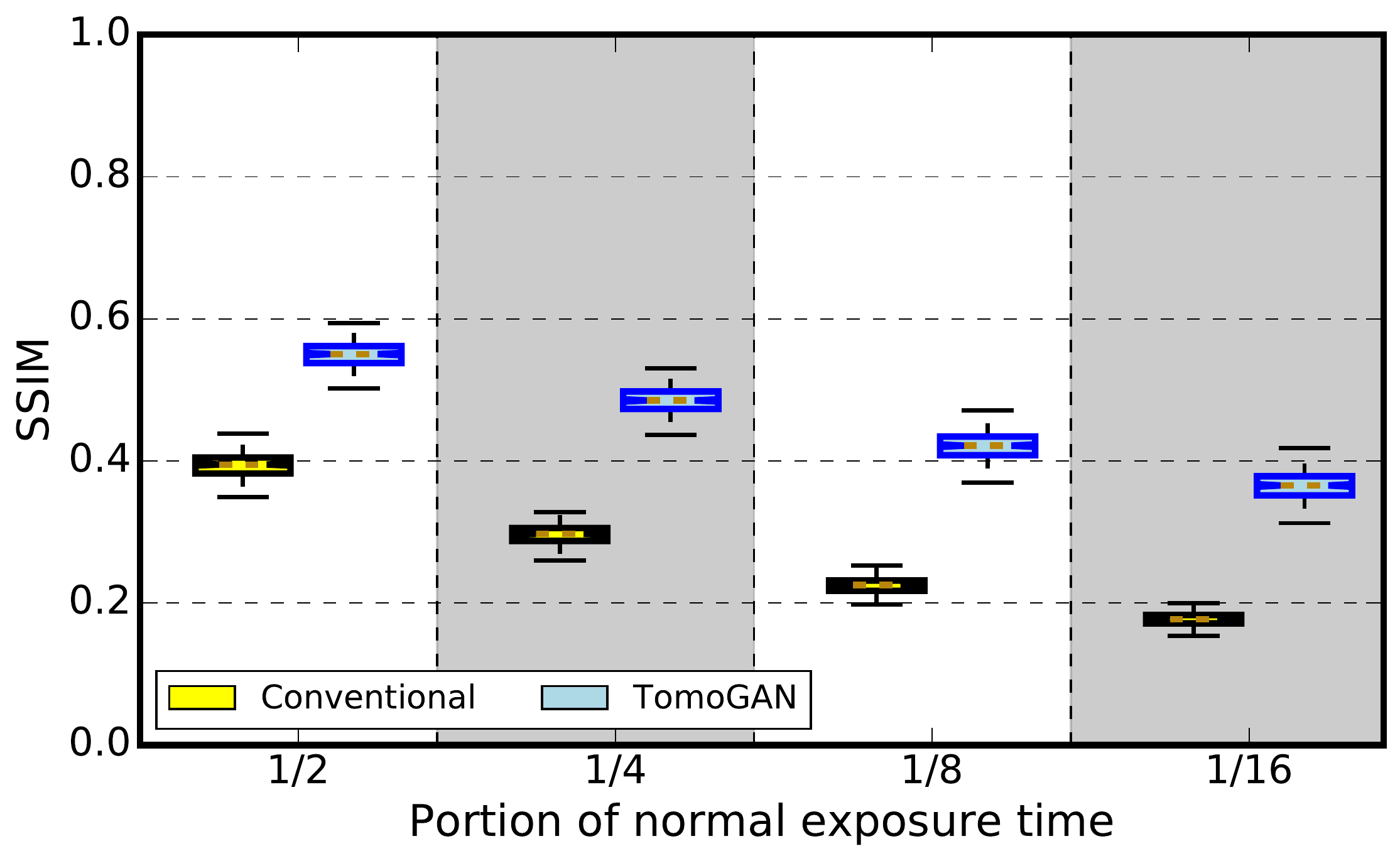}
\vspace{-4ex}
\caption{Shorter exposure time.}
\label{fig2:aps-ssim-exp}
\end{subfigure}
\caption{Per-slice SSIM similarities with a reconstruction of \num{1024}-projection experimental data, for both conventional and \TOMOGAN{}-enhanced reconstructions and for different degrees of subsampling. 
}
\label{fig:aps-ssim}
\end{figure}

The qualities of the conventional and \TOMOGAN{}-enhanced reconstructions using SSIM are shown in \autoref{fig2:aps-ssim-exp}.
Comparing \autoref{fig:aps-ssim} with \autoref{fig:simu-ssim}, we see that the improvement obtained for the experimental dataset is less than that obtained for the simulated dataset.
The features in the experimental dataset are much more diverse (different shapes, sizes, and contrast) compared with those of the simulated dataset (only circles), and therefore the denoising performance is less for the experimental datasets.

\begin{figure}[ht]
\centering
\begin{subfigure}{.48\linewidth}
\centering
\includegraphics[width=\textwidth]{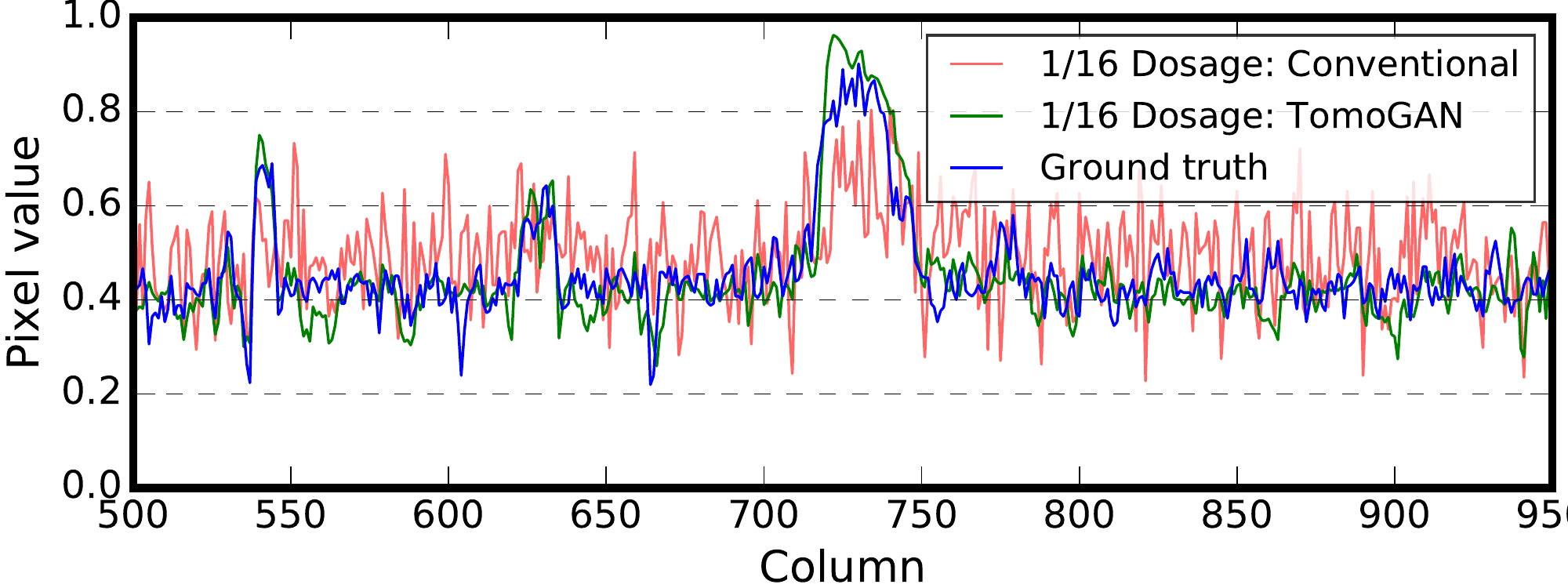}

\vspace{-0.5ex}

\caption*{$DS_{APS}^{\emph{B1}}$ .}
\end{subfigure}
\hspace{1ex}
\begin{subfigure}{.48\linewidth}
\centering
\includegraphics[width=\textwidth]{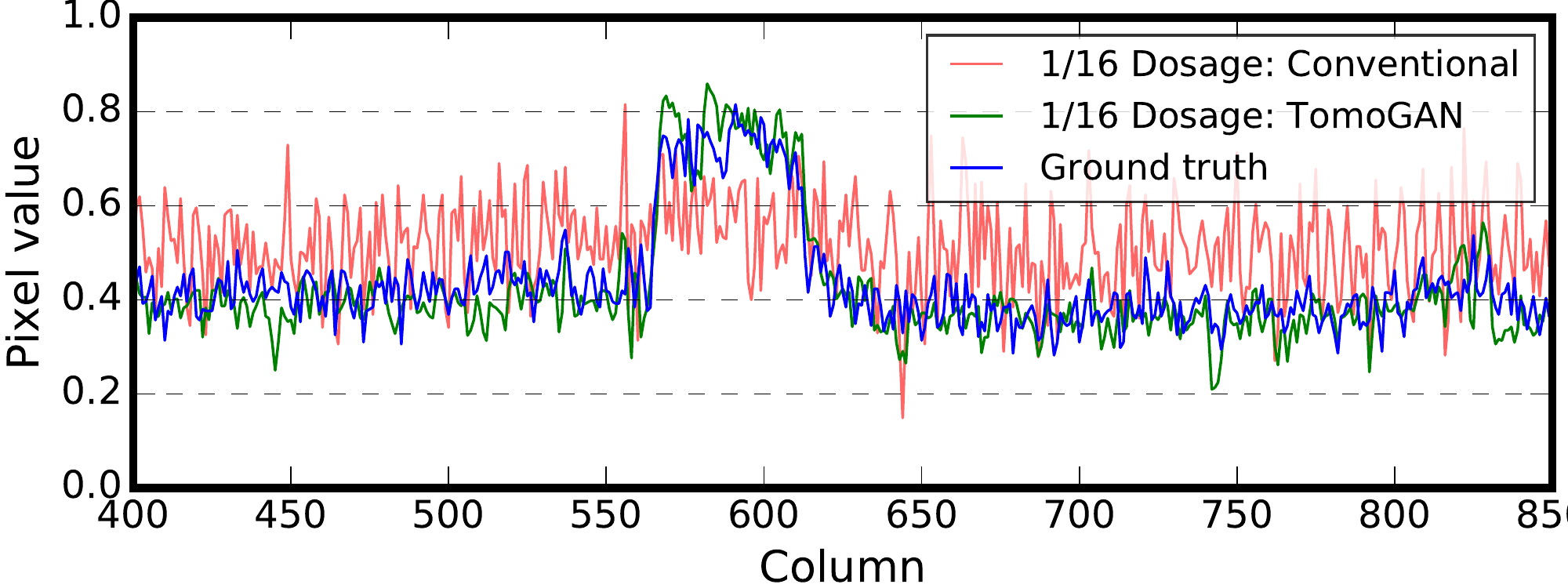}

\vspace{-0.5ex}

\caption*{$DS_{APS}^{\emph{N1}}$.}
\end{subfigure}

\vspace{1ex}

\begin{subfigure}{.48\linewidth}
\centering
\includegraphics[width=\textwidth]{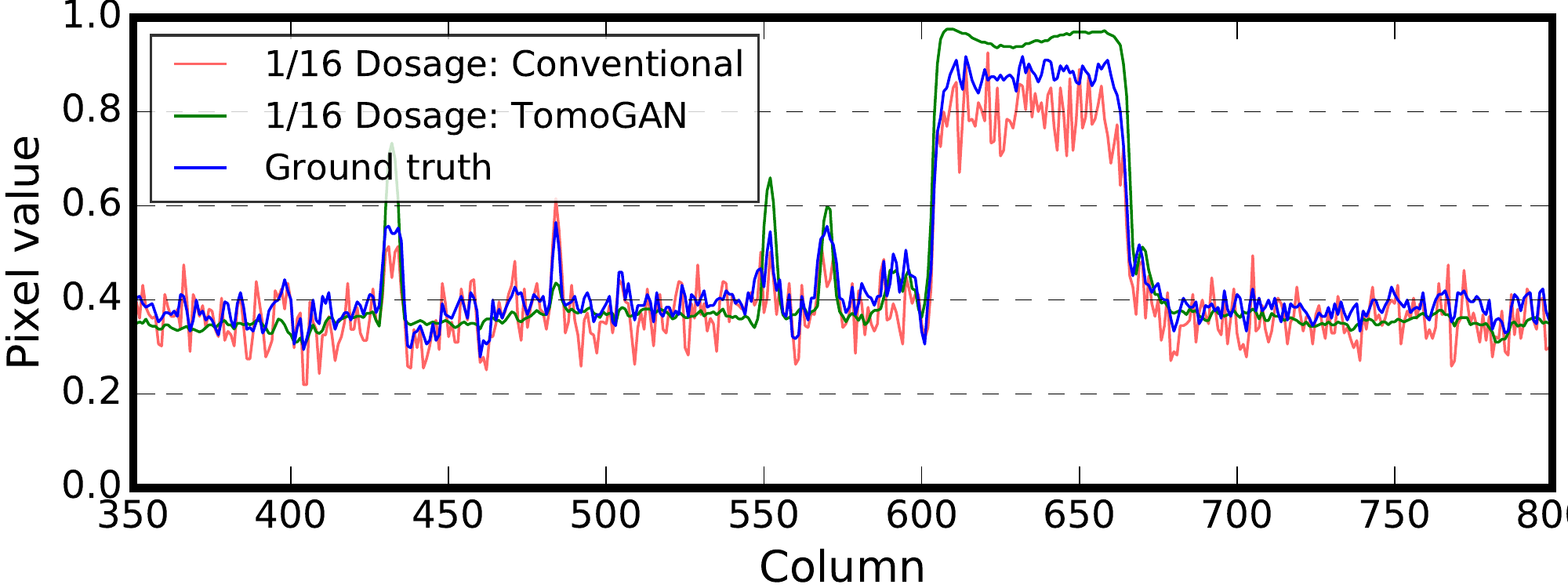}

\vspace{-0.5ex}

\caption*{$DS_{SLS}^{\emph{B1}}$.}
\end{subfigure}
\hspace{1ex}
\begin{subfigure}{.48\linewidth}
\centering
\includegraphics[width=\textwidth]{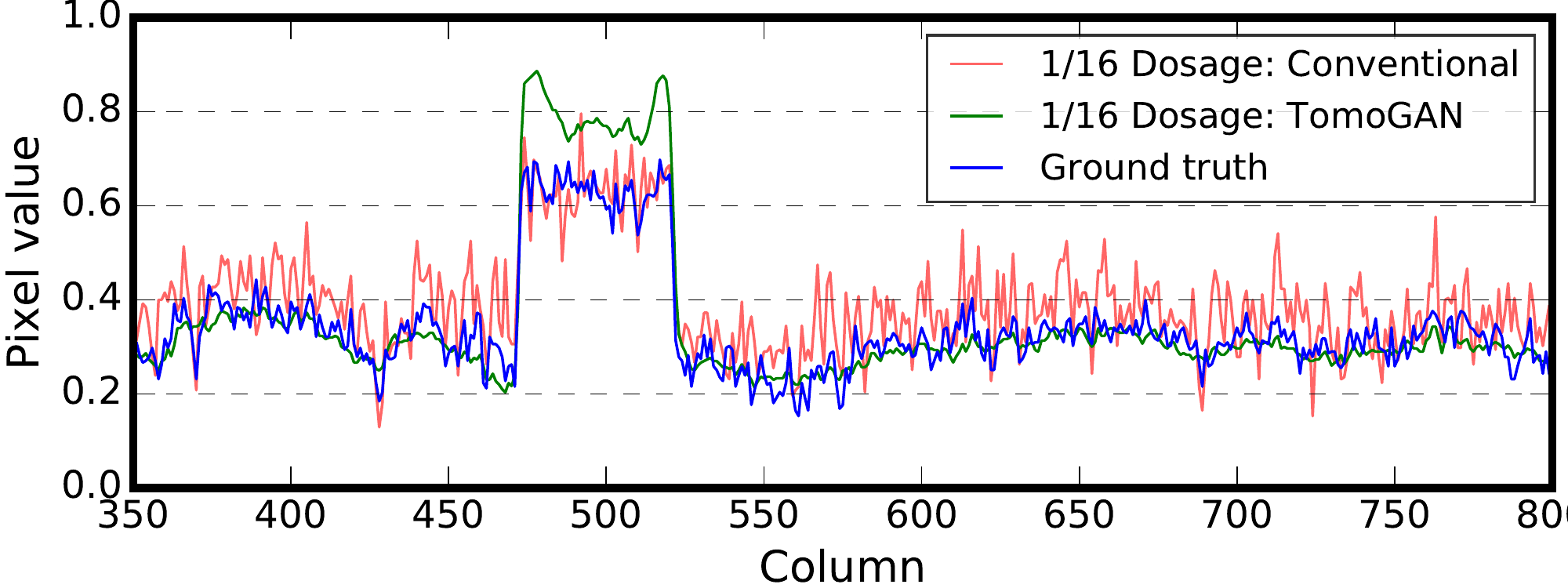}

\vspace{-0.5ex}

\caption*{$DS_{SLS}^{\emph{N1}}$.}
\end{subfigure}
\caption{Pixel values of an arbitrarily chosen feature in each of the four experimental datasets, with projections generated by using 1/16 of the normal exposure time. Feature shapes are different for each dataset.}
\label{fig:aps-all-ds-ept16-pixel} 
\end{figure}

We also show in \autoref{fig:aps-all-ds-ept16-pixel},  for the most challenging cases that use only 1/16 of normal exposure time to reconstruct,  the pixel value of a horizontal line that cross an arbitrarily chosen feature in each of the four datasets.

\subsection{Comparison with iterative methods}
\begin{figure}[ht] 
\centering
\begin{subfigure}{0.48\linewidth}
\centering 
\includegraphics[width=\textwidth,trim={3cm 8cm 0cm 0cm},clip]{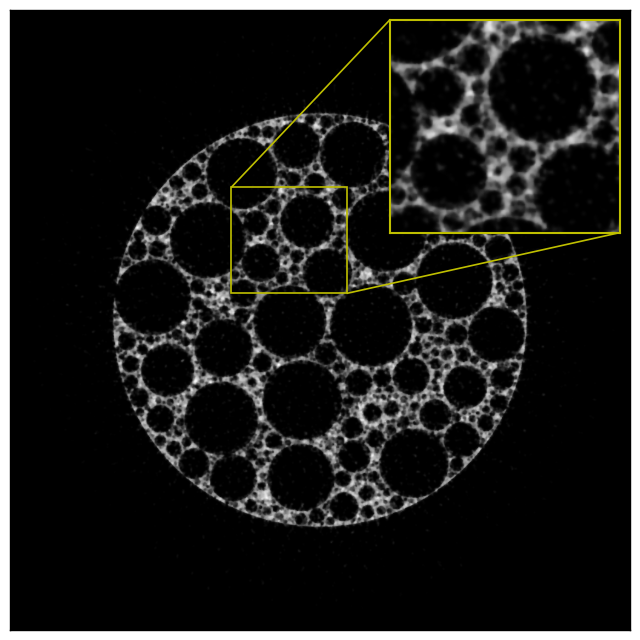}
\caption*{SIRT + total variation postprocess. }
\end{subfigure}
\begin{subfigure}{0.48\linewidth}
\centering
\includegraphics[width=\textwidth,trim={3cm 8cm 0cm 0cm},clip]{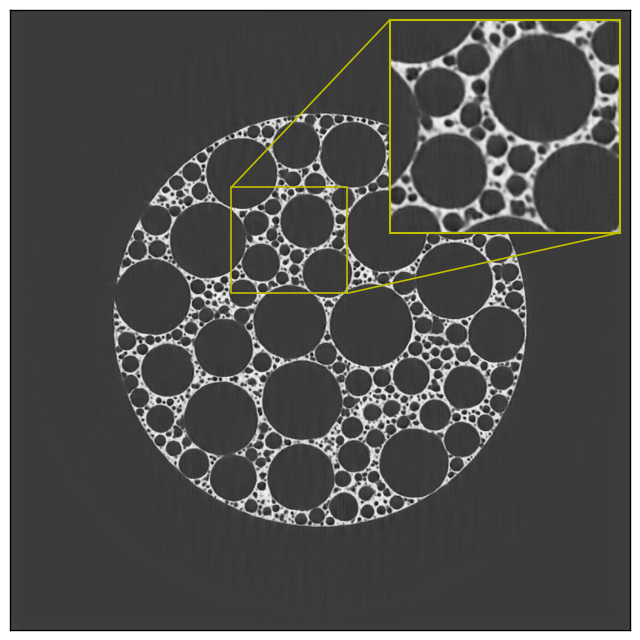}
\caption*{Filtered back projection + \TOMOGAN{} post-process.}
\end{subfigure}
\caption{SIRT + total variation vs.\ \TOMOGAN{}: an image reconstructed from 64 simulated projections.}
\label{fig:iter-vs-fbp+gan}
\end{figure}

Iterative methods are more resilient to noise in projections~\cite{geyer2015state}, but they are computationally demanding and prohibitively expensive for large data~\cite{Schindera2013, Bicer2017, Wang2017, bicer2015rapid}.
The filtered back-projection algorithm takes 40 ms to reconstruct one slice (using the TomoPy~\cite{gursoy2014tomopy,Pelt:pp5084} toolkit), and \TOMOGAN{} takes 30 ms to enhance the reconstruction, for a total of about 70 ms per slice. 
In contrast, the simultaneous iterative reconstruction technique (SIRT) takes 550 ms to reconstruct one slice with 400 iterations. 
These times are all measured on one NVIDIA Tesla V100-SXM2 graphic card.
Moreover, as illustrated in \autoref{fig:iter-vs-fbp+gan}, iterative reconstruction does not provide better image quality than that of our method. 

\subsection{Comparison with other solutions on experimental datasets}
Jin et al.~\cite{fbpconv} proposed a deep convolutional neural network (a variation of U-Net~\cite{unet} named FBPConvNet) for the sparse view computed tomography problem. 
Their network outperforms total variation-regularized iterative reconstruction for more realistic phantoms, and also runs faster. 
FBPConvNet, like \TOMOGAN{}'s generator, uses the FBP algorithm to reconstruct sparse X-ray projections, after which the CNN combines multi-resolution decomposition and residual learning in order to remove artifacts while preserving image structure.

FBPConvNet and \TOMOGAN{} differ in two key areas. 
First, \TOMOGAN{} uses a GAN architecture, with adversarial loss, to help train the generator, and a pre-trained VGG~\cite{vgg} network; results presented below suggest that the adversarial loss avoids artifacts.
Second, our generator, although also based on U-Net, has three U-Net boxes, as shown in \autoref{fig:unet} instead of the four in FBPConvNet, and no batch normalization layer~\cite{BatchNorm}, two factors that reduce computation and memory needs for inference (e.g., \TOMOGAN{} can efficiently run on Google edge TPU~\cite{abeykoon2019scientific}). 
For example, with one NVIDIA Tesla V100 card and a batch size of eight (minimizing DL framework overheads), \TOMOGAN{} and FBPConvNet take an average of 30ms and 90ms to process one 1024$\times$1024 image, respectively: \TOMOGAN{} is three times faster. 

Yang et al.~\cite{8340157} use Wasserstein GAN~\cite{2017arXiv170107875A} plus perceptual loss to denoise low-dose medical CT images,
with perceptual loss calculated with a pre-trained VGG~\cite{vgg} neural network. 
Their model, which we refer to as WGAN-VGG here,
differs from \TOMOGAN{} in two major respects.
First, WGAN-VGG uses different generator and discriminator architectures: while \TOMOGAN{} uses a variant of the commonly used U-net architecture as its generator, WGAN-VGG's is handcrafted. Second, \TOMOGAN{} incorporates information from adjacent slices to enhance images. Results such as those shown in \autoref{fig:depth-eva-simi} suggest that this use of multiple layers is beneficial in practice.

\begin{figure}[htbp]
\centering
\includegraphics[width=0.8\linewidth]{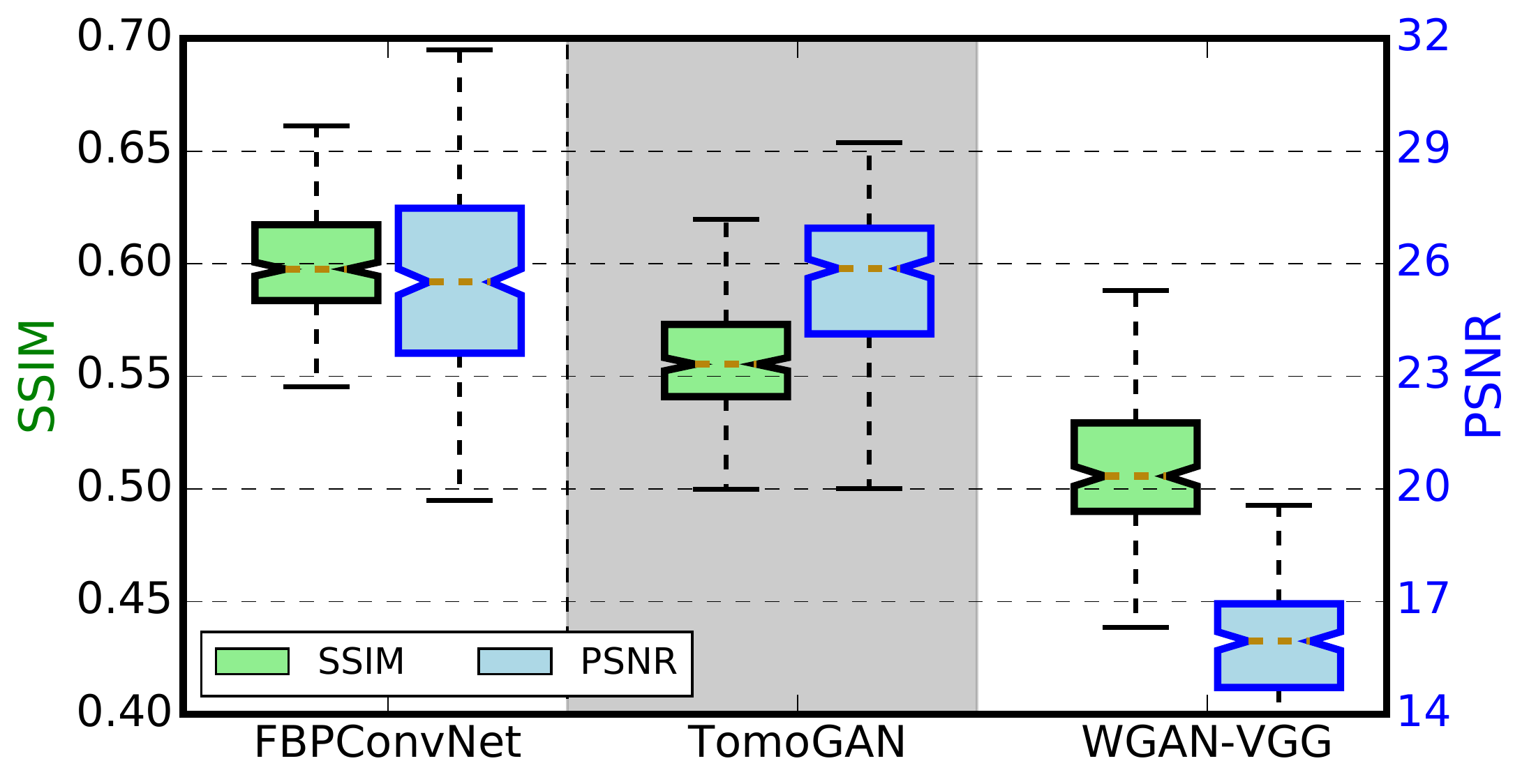}
\caption{SSIM and PSNR comparison between FBPConvNet, \TOMOGAN{}, and WGAN-VGG for a shale dataset.} 
\label{fig:fbpconvnet-cmp-simi} 
\end{figure}

\begin{figure*}[htbp]
\centering
\includegraphics[width=\textwidth]{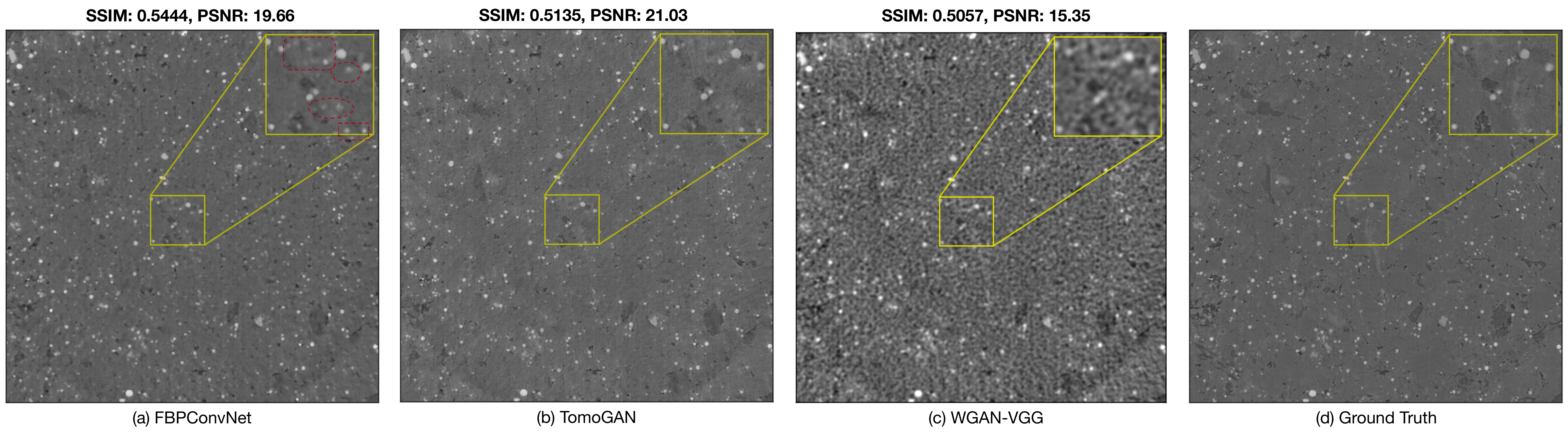}
\caption{Denoising performance of FBPConvNet, \TOMOGAN{}, and WGAN-VGG on an arbitrary slice of a shale sample.
} 
\label{fig:fbpconvnet-cmp-shale} 
\end{figure*}

To permit quantitative comparison of \TOMOGAN{} with FBPConvNet~\cite{fbpconv} and WGAN-VGG~\cite{8340157},
we downloaded the open source implementations of both the latter models and trained them with our experimental datasets. 
Specifically, we trained both models with $DS_{APS}^{\emph{B1}}$ and tested on $DS_{SLS}^{\emph{N1}}$.
\autoref{fig:fbpconvnet-cmp-simi} compares the SSIM and PSNR of FBPConvNet, \TOMOGAN{}, and WGAN-VGG.
\TOMOGAN{} outperforms FBPConvNet on PSNR (higher median and smaller variance means more stable performance) but has a worse SSIM score.
However, when we examine an arbitrarily chosen denoised image as shown in \autoref{fig:fbpconvnet-cmp-shale}, we see that FBPConvNet introduces low-contrast artifacts as marked in \autoref{fig:fbpconvnet-cmp-shale}(a) (there are visibly more white dots in \autoref{fig:fbpconvnet-cmp-shale}(a) than in \autoref{fig:fbpconvnet-cmp-shale}(d)),
likely because: (1) FBPConvNet only used MSE loss and those artifacts were not significant to the MSE, and/or (2) as observed by the FBPConvNet authors\cite{jimaging4110128}, FBPConvNet has a high risk of overfitting. 
But these artifacts are significant to \TOMOGAN{}'s adversarial loss and perceptual loss. 
WGAN-VGG has less good SSIM and PSNR values and,
as shown in \autoref{fig:fbpconvnet-cmp-shale}(c), does not work well for this dataset.
We attribute this failure to the fact that the model and parameters are tuned for brain image datasets, which have quite different characteristics to those of our hard X-ray shale sample dataset, e.g., synchrotron hard x-ray images have much more high-frequency content.
A model designed for one typically does not work well for the other.
So, we note that results in \autoref{fig:fbpconvnet-cmp-shale}(c) is not a fair comparison to WGAN-VGG.
However, we believe that this is not a well-understood fact in the literature, which furthermore includes much more work on deep learning for medical images. 
Thus, we view this comparison as a useful contribution, in that it shows that the medical image method cannot be used unchanged. 

\subsection{Model interpretation}
\begin{figure*}[ht] 
\centering
\includegraphics[width=\textwidth]{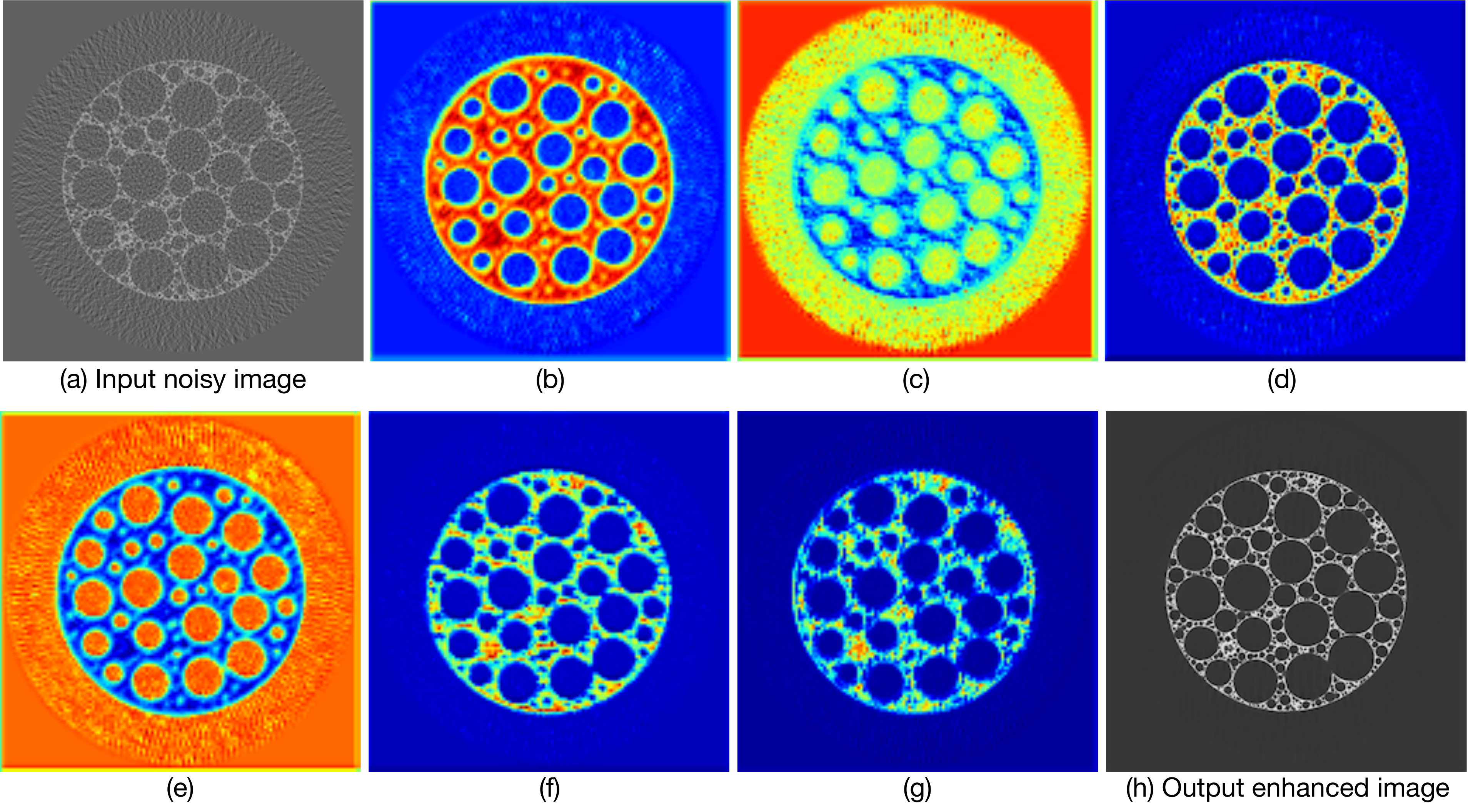}
\caption{Feature maps from the \TOMOGAN{} network (\autoref{fig:unet}): (a) input (noisy) image;
(b)--(g) six representative channels from the bottom-most layer; (h) final processed image.}
\label{fig:activation-map}
\end{figure*}

A straightforward technique for understanding the working of a deep convolutional neural network is to examine the outputs of its various layers---its \emph{feature maps}\cite{Zhou_2016}---during the forward pass. 
Recall from \autoref{fig:unet} that down-sampling reduces the image size to 1/8 of the original and increases the number of channels from $d$ to 128. 
Successful training requires that the convolutional kernel retain the important features as the image size decreases. 
To qualitatively understand how the trained generator improves image quality, we show in \autoref{fig:activation-map} the feature maps for an input (noisy) image, six representative channels from the bottom-most layer (the last layer of down sampling; red here indicates larger values and thus greater feature importance), and the final processed image.
We focus on the bottom-most layer because we expect it to show the most refined or important features selected by the preceding convolution kernels. 

Examining \autoref{fig:activation-map}(b)--(g) in detail, we see that (b) tends to keep dense regions of the sample (white in (h)), while (e) detects non-dense regions (air inside sample, black area in (h)); 
(c) detects the outer area in which there is no sample (should ideally be 0 in the output image); and
(d) is similar to (b) but pays more attention to the details. 
Studying (h) closely, or (even better) examining \autoref{fig:teaser-simu}, which shows a zoomed-in version of the same image, we see that some areas contain many small features in the form of small holes.
These small features could easily be lost in noise; interestingly, (f) and (g) seem to pay special attention to them. 
Intuitively, then, we may conclude that different channels (i.e., different convolution kernels) are responsible for capturing different types of features.

\section{Discussion}\label{sec:discuss}
We applied a generative adversarial network to improve the quality of images reconstructed from noisy x-ray tomographic projections. 
Experiments show that our model, \TOMOGAN{}, once trained on one sample,  can then be applied effectively to other similar samples, even if those samples are collected at a different facility and show different noise characteristics. 
Our results show that \TOMOGAN{} is general enough to provide visible improvement to images reconstructed from noisy projection datasets.
We claim that this method has great potential for studying samples with dynamic features because it allows for good-quality reconstructions from experiments that run much faster than conventional experiments,
either by collecting fewer projections or by using shorter exposure times per projection.

\section*{Acknowledgements}
\noindent
This work was supported in part by the U.S. Department of Energy, Office of Science, under contract DE-AC02-06CH11357. This research used resources of the Advanced Photon Source, a U.S. Department of Energy (DOE) Office of Science User Facility operated for the DOE Office of Science by Argonne National Laboratory under Contract No. DE-AC02-06CH11357. We thank Daniel M. Pelt at Centrum Wiskunde \& Informatica for sharing the simulated x-ray projection datasets.
We gratefully acknowledge the computing resources provided and operated by the Joint Laboratory for System Evaluation at Argonne National Laboratory.

\bibliographystyle{IEEEtran}
\bibliography{sample}

\begin{thebibliography}{10}
\providecommand{\url}[1]{#1}
\csname url@samestyle\endcsname
\providecommand{\newblock}{\relax}
\providecommand{\bibinfo}[2]{#2}
\providecommand{\BIBentrySTDinterwordspacing}{\spaceskip=0pt\relax}
\providecommand{\BIBentryALTinterwordstretchfactor}{4}
\providecommand{\BIBentryALTinterwordspacing}{\spaceskip=\fontdimen2\font plus
\BIBentryALTinterwordstretchfactor\fontdimen3\font minus
  \fontdimen4\font\relax}
\providecommand{\BIBforeignlanguage}[2]{{%
\expandafter\ifx\csname l@#1\endcsname\relax
\typeout{** WARNING: IEEEtran.bst: No hyphenation pattern has been}%
\typeout{** loaded for the language `#1'. Using the pattern for}%
\typeout{** the default language instead.}%
\else
\language=\csname l@#1\endcsname
\fi
#2}}
\providecommand{\BIBdecl}{\relax}
\BIBdecl

\bibitem{bonse1996x}
U.~Bonse and F.~Busch, ``X-ray computed microtomography ($\mu${CT}) using
  synchrotron radiation ({SR}),'' \emph{Progress in Biophysics and Molecular
  Biology}, vol.~65, no. 1-2, pp. 133--169, 1996.

\bibitem{wang2005sinogram}
J.~Wang, H.~Lu, T.~Li, and Z.~Liang, ``Sinogram noise reduction for low-dose ct
  by statistics-based nonlinear filters,'' in \emph{Medical Imaging 2005: Image
  Processing}, vol. 5747.\hskip 1em plus 0.5em minus 0.4em\relax International
  Society for Optics and Photonics, 2005, pp. 2058--2067.

\bibitem{wang2006penalized}
J.~Wang, T.~Li, H.~Lu, and Z.~Liang, ``Penalized weighted least-squares
  approach to sinogram noise reduction and image reconstruction for low-dose
  x-ray computed tomography,'' \emph{IEEE transactions on medical imaging},
  vol.~25, no.~10, pp. 1272--1283, 2006.

\bibitem{manduca2009projection}
A.~Manduca, L.~Yu, J.~D. Trzasko, N.~Khaylova, J.~M. Kofler, C.~M. McCollough,
  and J.~G. Fletcher, ``Projection space denoising with bilateral filtering and
  ct noise modeling for dose reduction in ct,'' \emph{Medical physics},
  vol.~36, no.~11, pp. 4911--4919, 2009.

\bibitem{Anirudh_2018_CVPR}
R.~Anirudh, H.~Kim, J.~J. Thiagarajan, K.~Aditya~Mohan, K.~Champley, and
  T.~Bremer, ``Lose the views: Limited angle ct reconstruction via implicit
  sinogram completion,'' in \emph{The IEEE Conference on Computer Vision and
  Pattern Recognition (CVPR)}, June 2018.

\bibitem{vogel1996iterative}
C.~R. Vogel and M.~E. Oman, ``Iterative methods for total variation
  denoising,'' \emph{SIAM Journal on Scientific Computing}, vol.~17, no.~1, pp.
  227--238, 1996.

\bibitem{beister2012iterative}
M.~Beister, D.~Kolditz, and W.~A. Kalender, ``Iterative reconstruction methods
  in x-ray ct,'' \emph{Physica medica}, vol.~28, no.~2, pp. 94--108, 2012.

\bibitem{Bazrafkan2019DeepLB}
S.~Bazrafkan, V.~V. Nieuwenhove, J.~Soons, J.~D. Beenhouwer, and J.~Sijbers,
  ``Deep learning based computed tomography whys and wherefores,''
  \emph{ArXiv}, vol. abs/1904.03908, 2019.

\bibitem{8327873}
A.~{Hauptmann}, F.~{Lucka}, M.~{Betcke}, N.~{Huynh}, J.~{Adler}, B.~{Cox},
  P.~{Beard}, S.~{Ourselin}, and S.~{Arridge}, ``Model-based learning for
  accelerated, limited-view 3-d photoacoustic tomography,'' \emph{IEEE
  Transactions on Medical Imaging}, vol.~37, no.~6, pp. 1382--1393, June 2018.

\bibitem{adler2018deep}
J.~Adler and O.~{\"O}ktem, ``Deep bayesian inversion,'' \emph{arXiv preprint
  arXiv:1811.05910}, 2018.

\bibitem{ma2011low}
J.~Ma, J.~Huang, Q.~Feng, H.~Zhang, H.~Lu, Z.~Liang, and W.~Chen, ``Low-dose
  computed tomography image restoration using previous normal-dose scan,''
  \emph{Medical physics}, vol.~38, no.~10, pp. 5713--5731, 2011.

\bibitem{8332971}
E.~Kang, W.~Chang, J.~Yoo, and J.~C. Ye, ``Deep convolutional framelet
  denoising for low-dose {CT} via wavelet residual network,'' \emph{IEEE
  Transactions on Medical Imaging}, vol.~37, no.~6, pp. 1358--1369, June 2018.

\bibitem{wolterink2017generative}
J.~M. Wolterink, T.~Leiner, M.~A. Viergever, and I.~I{\v{s}}gum, ``Generative
  adversarial networks for noise reduction in low-dose ct,'' \emph{IEEE
  transactions on medical imaging}, vol.~36, no.~12, pp. 2536--2545, 2017.

\bibitem{deep-img}
G.~Wang, ``A perspective on deep imaging,'' \emph{IEEE Access}, vol.~4, pp.
  8914--8924, 2016.

\bibitem{jimaging4110128}
\BIBentryALTinterwordspacing
D.~M. Pelt, K.~J. Batenburg, and J.~A. Sethian, ``Improving tomographic
  reconstruction from limited data using mixed-scale dense convolutional neural
  networks,'' \emph{Journal of Imaging}, vol.~4, no.~11, 2018. [Online].
  Available: \url{http://www.mdpi.com/2313-433X/4/11/128}
\BIBentrySTDinterwordspacing

\bibitem{2016arXiv160806993H}
G.~{Huang}, Z.~{Liu}, L.~{van der Maaten}, and K.~Q. {Weinberger}, ``Densely
  connected convolutional networks,'' \emph{ArXiv e-prints}, Aug. 2016.

\bibitem{Yang2018}
\BIBentryALTinterwordspacing
X.~Yang, V.~{De Andrade}, W.~Scullin, E.~L. Dyer, N.~Kasthuri, F.~{De Carlo},
  and D.~G{\"{u}}rsoy, ``Low-dose x-ray tomography through a deep convolutional
  neural network,'' \emph{Scientific Reports}, vol.~8, no.~1, p. 2575, 2018.
  [Online]. Available: \url{https://doi.org/10.1038/s41598-018-19426-7}
\BIBentrySTDinterwordspacing

\bibitem{LeCun2015}
Y.~LeCun, Y.~Bengio, and G.~Hinton, ``{Deep learning},'' \emph{Nature}, vol.
  521, p. 436, may 2015.

\bibitem{AlexNet}
\BIBentryALTinterwordspacing
A.~Krizhevsky, I.~Sutskever, and G.~E. Hinton, ``{ImageNet} classification with
  deep convolutional neural networks,'' \emph{Communications of the ACM},
  vol.~60, no.~6, pp. 84--90, May 2017. [Online]. Available:
  \url{http://doi.acm.org/10.1145/3065386}
\BIBentrySTDinterwordspacing

\bibitem{BatchNorm}
\BIBentryALTinterwordspacing
S.~Ioffe and C.~Szegedy, ``Batch normalization: Accelerating deep network
  training by reducing internal covariate shift,'' in \emph{32nd International
  Conference on Machine Learning}, vol.~37.\hskip 1em plus 0.5em minus
  0.4em\relax JMLR.org, 2015, pp. 448--456. [Online]. Available:
  \url{http://dl.acm.org/citation.cfm?id=3045118.3045167}
\BIBentrySTDinterwordspacing

\bibitem{Dropout}
\BIBentryALTinterwordspacing
N.~Srivastava, G.~Hinton, A.~Krizhevsky, I.~Sutskever, and R.~Salakhutdinov,
  ``Dropout: A simple way to prevent neural networks from overfitting,''
  \emph{Journal of Machine Learning Research}, vol.~15, no.~1, pp. 1929--1958,
  Jan. 2014. [Online]. Available:
  \url{http://dl.acm.org/citation.cfm?id=2627435.2670313}
\BIBentrySTDinterwordspacing

\bibitem{residual}
K.~He, X.~Zhang, S.~Ren, and J.~Sun, ``Deep residual learning for image
  recognition,'' in \emph{IEEE Conference on Computer Vision and Pattern
  Recognition}, June 2016, pp. 770--778.

\bibitem{sr-survey-19}
\BIBentryALTinterwordspacing
S.~Anwar, S.~Khan, and N.~Barnes, ``A deep journey into super-resolution: {A}
  survey,'' \emph{CoRR}, vol. abs/1904.07523, 2019. [Online]. Available:
  \url{http://arxiv.org/abs/1904.07523}
\BIBentrySTDinterwordspacing

\bibitem{2017arXiv171110925U}
D.~{Ulyanov}, A.~{Vedaldi}, and V.~{Lempitsky}, ``Deep image prior,''
  \emph{ArXiv e-prints}, Nov. 2017.

\bibitem{srgan}
C.~Ledig, L.~Theis, F.~Huszár, J.~Caballero, A.~Cunningham, A.~Acosta,
  A.~Aitken, A.~Tejani, J.~Totz, Z.~Wang, and W.~Shi, ``Photo-realistic single
  image super-resolution using a generative adversarial network,'' in
  \emph{IEEE Conference on Computer Vision and Pattern Recognition}, July 2017,
  pp. 105--114.

\bibitem{pix2pix2017}
P.~Isola, J.-Y. Zhu, T.~Zhou, and A.~A. Efros, ``Image-to-image translation
  with conditional adversarial networks,'' in \emph{IEEE Conference on Computer
  Vision and Pattern Recognition}, 2017, pp. 5967--5976.

\bibitem{2014arXiv1406.2661G}
I.~J. {Goodfellow}, J.~{Pouget-Abadie}, M.~{Mirza}, B.~{Xu}, D.~{Warde-Farley},
  S.~{Ozair}, A.~{Courville}, and Y.~{Bengio}, ``Generative adversarial
  networks,'' in \emph{Advances in Neural Information Processing Systems},
  2014, pp. 2672--2680.

\bibitem{8353466}
H.~Shan, Y.~Zhang, Q.~Yang, U.~Kruger, M.~K. Kalra, L.~Sun, W.~Cong, and
  G.~Wang, ``{3-D} convolutional encoder-decoder network for low-dose {CT} via
  transfer learning from a {2-D} trained network,'' \emph{IEEE Transactions on
  Medical Imaging}, vol.~37, no.~6, pp. 1522--1534, June 2018.

\bibitem{8340157}
Q.~Yang, P.~Yan, Y.~Zhang, H.~Yu, Y.~Shi, X.~Mou, M.~K. Kalra, Y.~Zhang,
  L.~Sun, and G.~Wang, ``Low-dose {CT} image denoising using a generative
  adversarial network with {W}asserstein distance and perceptual loss,''
  \emph{IEEE Transactions on Medical Imaging}, vol.~37, no.~6, pp. 1348--1357,
  June 2018.

\bibitem{abs-1708-08333}
\BIBentryALTinterwordspacing
Y.~Han and J.~C. Ye, ``Framing u-net via deep convolutional framelets:
  Application to sparse-view {CT},'' \emph{CoRR}, vol. abs/1708.08333, 2017.
  [Online]. Available: \url{http://arxiv.org/abs/1708.08333}
\BIBentrySTDinterwordspacing

\bibitem{Du:18}
\BIBentryALTinterwordspacing
M.~Du, R.~Vescovi, K.~Fezzaa, C.~Jacobsen, and D.~G\"{u}rsoy, ``X-ray
  tomography of extended objects: a comparison of data acquisition
  approaches,'' \emph{J. Opt. Soc. Am. A}, vol.~35, no.~11, pp. 1871--1879, Nov
  2018. [Online]. Available:
  \url{http://josaa.osa.org/abstract.cfm?URI=josaa-35-11-1871}
\BIBentrySTDinterwordspacing

\bibitem{ssim}
Z.~Wang, A.~C. Bovik, H.~R. Sheikh, and E.~P. Simoncelli, ``Image quality
  assessment: From error visibility to structural similarity,'' \emph{IEEE
  Transactions on Image Processing}, vol.~13, no.~4, pp. 600--612, April 2004.

\bibitem{Ching:2017:xdesign}
\BIBentryALTinterwordspacing
D.~J. Ching and D.~G{\"{u}}rsoy, ``{{\it XDesign}: an open-source software
  package for designing X-ray imaging phantoms and experiments},''
  \emph{Journal of Synchrotron Radiation}, vol.~24, no.~2, pp. 537--544, Mar
  2017. [Online]. Available: \url{https://doi.org/10.1107/S1600577517001928}
\BIBentrySTDinterwordspacing

\bibitem{scikit:2014}
S.~van~der Walt, J.~Schönberger, J.~Nunez-Iglesias, F.~Boulogne, J.~Warner,
  N.~Yager, E.~Gouillart, T.~Yu, and t.~scikit-image contributors,
  ``scikit-image: Image processing in python,'' \emph{PeerJ}, vol.~2, 07 2014.

\bibitem{2017arXiv170107875A}
M.~{Arjovsky}, S.~{Chintala}, and L.~{Bottou}, ``{Wasserstein GAN},''
  \emph{ArXiv e-prints}, Jan. 2017.

\bibitem{improved-wgan}
\BIBentryALTinterwordspacing
I.~Gulrajani, F.~Ahmed, M.~Arjovsky, V.~Dumoulin, and A.~C. Courville,
  ``Improved training of {W}asserstein {GANs},'' \emph{CoRR}, 2017. [Online].
  Available: \url{http://arxiv.org/abs/1704.00028}
\BIBentrySTDinterwordspacing

\bibitem{Nasrollahi2014}
\BIBentryALTinterwordspacing
K.~Nasrollahi and T.~B. Moeslund, ``Super-resolution: A comprehensive survey,''
  \emph{Machine Vision and Applications}, vol.~25, no.~6, pp. 1423--1468, Aug
  2014. [Online]. Available: \url{https://doi.org/10.1007/s00138-014-0623-4}
\BIBentrySTDinterwordspacing

\bibitem{Tibbs:18}
\BIBentryALTinterwordspacing
A.~B. Tibbs, I.~M. Daly, N.~W. Roberts, and D.~R. Bull, ``Denoising imaging
  polarimetry by adapted bm3d method,'' \emph{J. Opt. Soc. Am. A}, vol.~35,
  no.~4, pp. 690--701, Apr 2018. [Online]. Available:
  \url{http://josaa.osa.org/abstract.cfm?URI=josaa-35-4-690}
\BIBentrySTDinterwordspacing

\bibitem{Li:15}
\BIBentryALTinterwordspacing
C.~Li, Y.~Ma, J.~Huang, X.~Mei, and J.~Ma, ``Hyperspectral image denoising
  using the robust low-rank tensor recovery,'' \emph{J. Opt. Soc. Am. A},
  vol.~32, no.~9, pp. 1604--1612, Sep 2015. [Online]. Available:
  \url{http://josaa.osa.org/abstract.cfm?URI=josaa-32-9-1604}
\BIBentrySTDinterwordspacing

\bibitem{KimLL15a}
J.~Kim, J.~K. Lee, and K.~M. Lee, ``Deeply-recursive convolutional network for
  image super-resolution,'' in \emph{IEEE Conference on Computer Vision and
  Pattern Recognition}, June 2016, pp. 1637--1645.

\bibitem{loss-cmp}
\BIBentryALTinterwordspacing
H.~Zhao, O.~Gallo, I.~Frosio, and J.~Kautz, ``Loss functions for neural
  networks for image processing,'' \emph{CoRR}, vol. abs/1511.08861, 2015.
  [Online]. Available: \url{http://arxiv.org/abs/1511.08861}
\BIBentrySTDinterwordspacing

\bibitem{vgg}
\BIBentryALTinterwordspacing
K.~Simonyan and A.~Zisserman, ``Very deep convolutional networks for
  large-scale image recognition,'' \emph{CoRR}, vol. abs/1409.1556, 2014.
  [Online]. Available: \url{http://arxiv.org/abs/1409.1556}
\BIBentrySTDinterwordspacing

\bibitem{dlhub18}
\BIBentryALTinterwordspacing
R.~Chard, Z.~Li, K.~Chard, L.~T. Ward, Y.~N. Babuji, A.~Woodard, S.~Tuecke,
  B.~Blaiszik, M.~J. Franklin, and I.~T. Foster, ``Dlhub: Model and data
  serving for science,'' \emph{CoRR}, vol. abs/1811.11213, 2018. [Online].
  Available: \url{http://arxiv.org/abs/1811.11213}
\BIBentrySTDinterwordspacing

\bibitem{B19}
W.~Kanitpanyacharoen, D.~Parkinson, F.~De~Carlo, F.~Marone, M.~Stampanoni,
  R.~Mokso, A.~MacDowell, and H.-R. Wenk, \emph{The Tomography Round-Robin
  datasets}.

\bibitem{B18}
\BIBentryALTinterwordspacing
W.~Kanitpanyacharoen, D.~Y. Parkinson, F.~De~Carlo, F.~Marone, M.~Stampanoni,
  R.~Mokso, A.~MacDowell, and H.-R. Wenk, ``{A comparative study of X-ray
  tomographic microscopy on shales at different synchrotron facilities: ALS,
  APS and SLS},'' \emph{Journal of Synchrotron Radiation}, vol.~20, no.~1, pp.
  172--180, Jan 2013. [Online]. Available:
  \url{https://doi.org/10.1107/S0909049512044354}
\BIBentrySTDinterwordspacing

\bibitem{DeCarlo:2018:tomobank}
\BIBentryALTinterwordspacing
F.~D. Carlo, D.~Gürsoy, D.~J. Ching, K.~J. Batenburg, W.~Ludwig, L.~Mancini,
  F.~Marone, R.~Mokso, D.~M. Pelt, J.~Sijbers, and M.~Rivers, ``{TomoBank}: a
  tomographic data repository for computational x-ray science,''
  \emph{Measurement Science and Technology}, vol.~29, no.~3, p. 034004, feb
  2018. [Online]. Available: \url{https://doi.org/10.1088%2F1361-6501%2Faa9c19}
\BIBentrySTDinterwordspacing

\bibitem{ASTRA}
\BIBentryALTinterwordspacing
W.~van Aarle, W.~J. Palenstijn, J.~D. Beenhouwer, T.~Altantzis, S.~Bals, K.~J.
  Batenburg, and J.~Sijbers, ``The {ASTRA Toolbox}: A platform for advanced
  algorithm development in electron tomography,'' \emph{Ultramicroscopy}, vol.
  157, pp. 35 -- 47, 2015. [Online]. Available:
  \url{http://www.sciencedirect.com/science/article/pii/S0304399115001060}
\BIBentrySTDinterwordspacing

\bibitem{gursoy2014tomopy}
D.~G{\"u}rsoy, F.~De~Carlo, X.~Xiao, and C.~Jacobsen, ``Tomopy: a framework for
  the analysis of synchrotron tomographic data,'' \emph{Journal of synchrotron
  radiation}, vol.~21, no.~5, pp. 1188--1193, 2014.

\bibitem{Pelt:pp5084}
\BIBentryALTinterwordspacing
D.~M. Pelt, D.~G{\"{u}}rsoy, W.~J. Palenstijn, J.~Sijbers, F.~De~Carlo, and
  K.~J. Batenburg, ``{Integration of {TomoPy} and the {ASTRA} toolbox for
  advanced processing and reconstruction of tomographic synchrotron data},''
  \emph{Journal of Synchrotron Radiation}, vol.~23, no.~3, pp. 842--849, May
  2016. [Online]. Available: \url{https://doi.org/10.1107/S1600577516005658}
\BIBentrySTDinterwordspacing

\bibitem{Reiffen:63}
B.~Reiffen, ``A note on "very noisy" channels,'' \emph{Information and
  Control}, vol.~6, pp. 126--130, 1963.

\bibitem{boas2012ct}
F.~E. Boas and D.~Fleischmann, ``{CT Artifacts: Causes and Reduction
  Techniques},'' \emph{Imaging Med}, vol.~4, no.~2, pp. 229--240, 2012.

\bibitem{abadi2016tensorflow}
\BIBentryALTinterwordspacing
M.~Abadi, P.~Barham, J.~Chen, Z.~Chen, A.~Davis, J.~Dean, M.~Devin,
  S.~Ghemawat, G.~Irving, M.~Isard, M.~Kudlur, J.~Levenberg, R.~Monga,
  S.~Moore, D.~G. Murray, B.~Steiner, P.~Tucker, V.~Vasudevan, P.~Warden,
  M.~Wicke, Y.~Yu, and X.~Zheng, ``{TensorFlow}: A system for large-scale
  machine learning,'' in \emph{12th USENIX Conference on Operating Systems
  Design and Implementation}, ser. OSDI'16.\hskip 1em plus 0.5em minus
  0.4em\relax Berkeley, CA, USA: USENIX Association, 2016, pp. 265--283.
  [Online]. Available: \url{http://dl.acm.org/citation.cfm?id=3026877.3026899}
\BIBentrySTDinterwordspacing

\bibitem{Adam}
\BIBentryALTinterwordspacing
D.~P. Kingma and J.~Ba, ``Adam: {A} method for stochastic optimization,''
  \emph{CoRR}, 2014. [Online]. Available: \url{http://arxiv.org/abs/1412.6980}
\BIBentrySTDinterwordspacing

\bibitem{fbpconv}
K.~H. {Jin}, M.~T. {McCann}, E.~{Froustey}, and M.~{Unser}, ``Deep
  convolutional neural network for inverse problems in imaging,'' \emph{IEEE
  Transactions on Image Processing}, vol.~26, no.~9, pp. 4509--4522, Sep. 2017.

\bibitem{liu2019deep}
Z.~Liu, T.~Bicer, R.~Kettimuthu, and I.~Foster, ``Deep learning accelerated
  light source experiments,'' \emph{arXiv preprint arXiv:1910.04081}, 2019.

\bibitem{geyer2015state}
L.~L. Geyer, U.~J. Schoepf, F.~G. Meinel, J.~W. Nance~Jr, G.~Bastarrika, J.~A.
  Leipsic, N.~S. Paul, M.~Rengo, A.~Laghi, and C.~N. De~Cecco, ``State of the
  art: iterative ct reconstruction techniques,'' \emph{Radiology}, vol. 276,
  no.~2, pp. 339--357, 2015.

\bibitem{Schindera2013}
S.~T. Schindera, D.~Odedra, S.~A. Raza, T.~K. Kim, H.-J. Jang, Z.~Szucs-Farkas,
  and P.~Rogalla, ``Iterative reconstruction algorithm for ct: Can radiation
  dose be decreased while low-contrast detectability is preserved?''
  \emph{Radiology}, vol. 269, no.~2, pp. 511--518, 2013, pMID: 23788715.

\bibitem{Bicer2017}
T.~Bicer, D.~G{\"u}rsoy, V.~D. Andrade, R.~Kettimuthu, W.~Scullin, F.~D. Carlo,
  and I.~T. Foster, ``Trace: a high-throughput tomographic reconstruction
  engine for large-scale datasets,'' \emph{Advanced Structural and Chemical
  Imaging}, vol.~3, no.~1, p.~6, Jan 2017.

\bibitem{Wang2017}
X.~Wang, A.~Sabne, P.~Sakdhnagool, S.~J. Kisner, C.~A. Bouman, and S.~P.
  Midkiff, ``Massively parallel 3d image reconstruction,'' in \emph{Proceedings
  of the International Conference for High Performance Computing, Networking,
  Storage and Analysis}, ser. SC '17.\hskip 1em plus 0.5em minus 0.4em\relax
  New York, NY, USA: ACM, 2017, pp. 3:1--3:12.

\bibitem{bicer2015rapid}
T.~Bicer, D.~Gursoy, R.~Kettimuthu, F.~De~Carlo, G.~Agrawal, and I.~T. Foster,
  ``Rapid tomographic image reconstruction via large-scale parallelization,''
  in \emph{European Conference on Parallel Processing}.\hskip 1em plus 0.5em
  minus 0.4em\relax Springer, 2015, pp. 289--302.

\bibitem{unet}
O.~Ronneberger, P.~Fischer, and T.~Brox, ``U-net: Convolutional networks for
  biomedical image segmentation,'' in \emph{Medical Image Computing and
  Computer-Assisted Intervention}, N.~Navab, J.~Hornegger, W.~M. Wells, and
  A.~F. Frangi, Eds.\hskip 1em plus 0.5em minus 0.4em\relax Cham: Springer
  International Publishing, 2015, pp. 234--241.

\bibitem{abeykoon2019scientific}
V.~Abeykoon, Z.~Liu, R.~Kettimuthu, G.~Fox, and I.~Foster, ``Scientific image
  restoration anywhere,'' \emph{arXiv preprint arXiv:1911.05878}, 2019.

\bibitem{Zhou_2016}
B.~Zhou, A.~Khosla, A.~Lapedriza, A.~Oliva, and A.~Torralba, ``Learning deep
  features for discriminative localization,'' \emph{2016 IEEE Conference on
  Computer Vision and Pattern Recognition (CVPR)}, Jun 2016.

\end{thebibliography}

\section*{License}
\noindent
The submitted manuscript has been created by UChicago Argonne, LLC, Operator of Argonne National Laboratory (``Argonne"). Argonne, a U.S. Department of Energy Office of Science laboratory, is operated under Contract No. DE-AC02-06CH11357. The U.S. Government retains for itself, and others acting on its behalf, a paid-up nonexclusive, irrevocable worldwide license in said article to reproduce, prepare derivative works, distribute copies to the public, and perform publicly and display publicly, by or on behalf of the Government. The Department of Energy will provide public access to these results of federally sponsored research in accordance with the DOE Public Access Plan. http://energy.gov/downloads/doe-public-access-plan.

\end{document}